\documentclass[journal]{IEEEtran}
\usepackage{amsmath,amsfonts, amsthm, amssymb}
\usepackage[linesnumbered,ruled,vlined]{algorithm2e}
\usepackage{array}
\usepackage[caption=false,font=footnotesize,labelfont=rm,textfont=rm]{subfig}
\usepackage{orcidlink}

\usepackage{textcomp}
\usepackage{stfloats}
\usepackage{url}
\usepackage{verbatim}
\usepackage{graphicx}
\usepackage{cite}
\usepackage{amssymb}
\usepackage{multirow}
\usepackage{booktabs}
\usepackage{amsthm,amssymb}
\usepackage{mathrsfs}
\usepackage{enumitem}
\usepackage{ulem}

\newtheorem{definition}{Definition}

\newtheorem{assumption}{Assumptions}

\begin{document}

\title{SacFL: Self-Adaptive Federated Continual Learning for Resource-Constrained End Devices}

\author{Zhengyi Zhong$^{\orcidlink{0000-0002-1515-4876}}$, Weidong Bao$^{\orcidlink{0000-0003-1867-3660}}$, Ji Wang$^{\orcidlink{0000-0002-4199-2793}}$, Jianguo Chen$^{\orcidlink{0000-0001-5009-578X}}$, Lingjuan Lyu$^{\orcidlink{0000-0003-3170-4994}}$, Wei Yang Bryan Lim$^{\orcidlink{0000-0003-2150-5561}}$

\thanks{Manuscript received May 28, 2024; accepted April 27, 2025. This work was partially funded by the National Natural Science Foundation of China under Grant 62372486, the Pearl River Talent Plan under Grant 2023QN10X579, and the Natural Science Foundation of Guangdong Province under Grant 2023A1515011179. (\textit{Corresponding author: Ji Wang}).

Zhengyi Zhong, Weidong Bao, Ji Wang are with the Laboratory for Big Data and Decision, National University of Defense Technology, Changsha 410073, China (e-mail: zhongzhengyi20@nudt.edu.cn; wdbao@nudt.edu.cn; wangji@nudt.edu.cn).
 
Jianguo Chen is with the School of Software Engineering, Sun Yat-sen University, China (e-mail: chenjg33@mail.sysu.edu.cn).

Lingjuan Lyu is with Sony AI, Japan (e-mail: lingjuanlvsmile@gmail.com). 

Wei Yang Bryan Lim is with Nanyang Technological University, Singapore (e-mail: bryan.limwy@ntu.edu.sg).}

}

\markboth{Journal of \LaTeX\ Class Files,~Vol.~14, No.~8, August~202X}%
{Shell \MakeLowercase{\textit{et al.}}: A Sample Article Using IEEEtran.cls for IEEE Journals}


\maketitle

\begin{abstract}
The proliferation of end devices has led to a distributed computing paradigm, wherein on-device machine learning models continuously process diverse data generated by these devices. The dynamic nature of this data, characterized by continuous changes or \textit{data drift}, poses significant challenges for on-device models. To address this issue, continual learning (CL) is proposed, enabling machine learning models to incrementally update their knowledge and mitigate \textit{catastrophic forgetting}. However, the traditional centralized approach to CL is unsuitable for end devices due to privacy and data volume concerns. In this context, federated continual learning (FCL) emerges as a promising solution, preserving user data locally while enhancing models through collaborative updates. Aiming at the challenges of limited storage resources for CL, poor autonomy in task shift detection, and difficulty in coping with new adversarial tasks in FCL scenario, we propose a novel FCL framework named $\rm{SacFL}$. $\rm{SacFL}$ employs an Encoder-Decoder architecture to separate task-robust and task-sensitive components, significantly reducing storage demands by retaining lightweight task-sensitive components for resource-constrained end devices. Moreover, $\rm{SacFL}$ leverages contrastive learning to introduce an autonomous data shift detection mechanism, enabling it to discern whether a new task has emerged and whether it is a benign task. This capability ultimately allows the device to autonomously trigger CL or attack defense strategy without additional information, which is more practical for end devices. Comprehensive experiments conducted on multiple text and image datasets, such as Cifar100 and THUCNews, have validated the effectiveness of $\rm{SacFL}$ in both class-incremental and domain-incremental scenarios. Furthermore, a demo system has been developed to verify its practicality.

\end{abstract}


\section{Introduction}
\IEEEPARstart{I}{n} recent years, the rapid development of end devices has given rise to a distributed intelligent computing paradigm. Within this framework, these devices generate vast amounts of data, including images, text, and audio. Over time, the collected data undergoes continuous changes, a phenomenon known as \textit{data drift}. Training a subsequent task with a model previously trained on a different task results in a significant decline in performance on the original task. This phenomenon is known as \textit{catastrophic forgetting} \cite{Ge2023}. One of the primary challenges in training machine learning models is to enhance their capacity for continual learning or to mitigate the rate of forgetting.

 The predominant approach in CL primarily focuses on centralized scenarios \cite{Douillard2020}, where user data generated by end devices is transmitted to a central node for training. However, this approach has become increasingly unsuitable for portable devices. On the one hand, user data is often highly privacy-sensitive, and directly transferring this data to remote servers poses a significant threat to user privacy \cite{Lim2020, Yang2019}. On the other hand, the effectiveness of models relies on extensive datasets, but the data volume available on individual end devices is inadequate to fully support the training of robust models. Therefore, in the context of distributed end devices, it is crucial to address how to enable multiple end devices to collaboratively learn continually while ensuring the privacy of client data. Federated learning (FL) \cite{McMahan2017} has emerged as a promising solution to these challenges. FL uploads model updates to remote servers while preserving users' data locally, thereby enhancing the learning process of models in a distributed manner. Building upon this premise, our study aims to explore continual learning methods based on federated learning.

Unlike the centralized approach, federated continual learning requires each end device to perform continual learning, which introduces three distinct challenges: 

\begin{itemize}
    \item \textbf{\textit{C1:}} Using conventional CL methods requires retaining entire or major segments of past models or preserving a large amount of historical data, imposing considerable storage demands on end devices. However, the inherent hardware limitations result in scarce storage resources, leading to a significant storage burden on resource-constrained end devices in FL.

    \item \textbf{\textit{C2:}} Conventional CL methods typically require external intervention to notify the model of task changes or data drift, lacking inherent mechanisms to detect data drift and adaptively adjust the learning process. This is impractical in distributed end device scenarios, where numerous autonomous devices, such as surveillance cameras, operate without external intervention, making conventional CL methods unsuitable.

    \item \textbf{\textit{C3:}} Conventional CL methods typically assume that new data is benign. However, in the context of FL, a distributed environment where data on end devices is uncontrollable, it is difficult to prevent malicious clients from introducing adversarial data during the CL process, which can potentially disrupt the global historical knowledge. Current methods cannot continuously monitor adversarial data and defend against such attacks.
\end{itemize}

To address above challenges, we design an Encoder-Decoder architecture that splits the model into a task-robust Encoder and a lightweight task-sensitive Decoder based on the variation of tasks. Only the Decoder is preserved for historical tasks, while the Encoder model is shared among tasks and clients. This approach not only facilitates knowledge transfer both in temporal and spatial dimensions but also effectively alleviates the resource burden to end devices. Meanwhile, inspired by contrastive learning, we compare the distances between the Encoders before and after updates to determine if data drift occurs. If the distance exceeds a certain threshold, it indicates data drift, triggering the CL mechanism and allowing end devices to update knowledge in a self-adaptive manner. These approaches avoid the need for extra information (\textit{e.g.,} task ID and data label) and facilitate \textbf{f}ederated \textbf{c}ontinual \textbf{l}earning with \textbf{s}elf-\textbf{a}daptive ability (\textbf{$\rm{SacFL}$}).
Furthermore, once task changes are monitored, we further consider whether the new tasks are benign or not. We propose adversarial task monitoring and defense methods, enabling clients to autonomously assess whether a new task is adversarial and take corresponding defense measures to mitigate the impact of the attack. This approach enhances the adaptability of clients in federated CL under adversarial environments.

In summary, the contributions are as follows:

\begin{itemize}
	\item Breaking the conventional assumption of centralized continual learning by proposing a federated continual learning method called $\rm{SacFL}$. This method effectively integrates knowledge from resource-constrained devices while simultaneously reducing the resource requirements of continual learning. 
	
	\item We introduce a data shift detection method that enables end devices to autonomously trigger the CL mechanism without relying on extra information or sharing data with the server. This innovation significantly enhances the self-adaptive capability of model training on end devices while safeguarding privacy.
	
	\item To address the potential issue of encountering new adversarial data during the CL, an adversarial task detection method and defense strategy are proposed, enhancing the adaptability of $\rm{SacFL}$ in adversarial environments.
	
	\item We validate the effectiveness of $\rm{SacFL}$ using multiple image and text datasets, including FashionMNIST, Cifar10, Cifar100, and THUCNews. Evaluations are performed in both class-incremental and domain-incremental scenarios. Additionally, we conduct experiments on a demo system, further confirming its superiority.
\end{itemize}

\section{Related work}
\subsection{Continual Learning}
Current continual learning methods can be divided into three main categories: Regularization-based Approach, Replay-based Approach, and Architecture-based Approach \cite{Ge2023}.

The Regularization-based Approach aims to balance the model performance between new and old tasks by adding regularization terms during the training process of new tasks, thus preventing catastrophic forgetting. Specifically, regularization can be applied at both the parameter and function level. At the parameter level, the importance of model parameters is computed to identify the parameters that contribute significantly to the computation results. Penalty regularization terms are then added to these parameters, allowing them to retain knowledge from old tasks \cite{Kirkpatrick2017}. In addition, freezing certain important parameters or reducing their learning rate can be regarded as variants of this regularization method. At the function level, knowledge distillation \cite{Gou2021} is commonly used to preserve old knowledge \cite{Li2017}. When complete data for the old tasks are not available, inference can be performed using incremental data, additional unlabeled data, or generated data \cite{Dhar2019}. Furthermore, when only partial data for previous tasks are accessible, data replay and knowledge distillation can be combined to enhance performance \cite{Douillard2020}.

The Replay-based Approach has three primary sub-directions: experience replay, generative replay, and feature replay \cite{Wang2023, li2024towards}. Experience replay involves constructing a replay buffer to store a small amount of historical data, which is then replayed during the training of subsequent tasks to enhance the model's learning ability \cite{Chaudhry2019}. In addition to experience replay, generative replay involves generating data using generative models. Instead of replaying old samples, generated data is used to retain memory throughout the continual learning process \cite{Boschini2022, Xiang2019}. Feature replay, on the other hand, replays the features of old data by utilizing feature extractors \cite{Liu2020, Gong2022}.

The above methods are based on parameter sharing between different tasks. In contrast, the Architecture-based Approach takes a different approach by implementing separate model structures for different tasks at the architectural level, achieving parameter isolation between tasks to avoid catastrophic forgetting. Typical methods include parameter allocation, model decomposition, and modular networks. Parameter allocation involves freezing key parameters for each task using masks, while the remaining parameters are used for training subsequent tasks \cite{Mallya2018, Xue2022}. Model decomposition decomposes the model into task-sharing and task-specific components, where the task-specific model expands as the number of tasks increases \cite{Mehta2021}. On the other hand, modular networks establish a subnetwork for each incremental task; however, this may incur significant memory overhead \cite{Jathushan2019}.

Currently, majority of CL methods are developed under the assumption that new data is reliable, and research on the robustness of CL is very limited \cite{Bai2023}. \cite{Bai2023} is the first to investigate the vulnerability of CL models to adversarial attacks.  It employs a replay-based approach, enhancing the robustness of CL by training on boundary samples selected from both old and new tasks. \cite{Khan2022} enhances resistance to adversarial attacks by training the model on robust features derived from the original data.  However, these methods are considered preemptive defenses.  This paper focuses on remedial measures, specifically how to identify the adversarial new samples during the CL and how to mitigate harms.

\subsection{Federated Learning}
Federated Learning was proposed by Google in 2016 \cite{McMahan2017} as a way to transfer model parameters instead of data, reducing the privacy leakage risk in traditional cloud computing. Federated learning can be categorized into three types: horizontal federated learning, vertical federated learning \cite{Lim2020}, and transfer federated learning \cite{Yang2019, FLSurveyandBenchmarkforGenRobFair_TPAMI24}. Horizontal federated learning is currently a research hotspot and focuses on several areas.

\begin{itemize}[leftmargin=*,noitemsep,topsep=0pt]
	\item Personalization:  Under the federated learning framework, clients' personalized demands can be categorized into data heterogeneity, system heterogeneity, and task heterogeneity \cite{Zhong2023, FCCLPlus_TPAMI23}. Techniques used in this area include Adding User Context \cite{Kairouz2021}, Meta-Learning \cite{Fallah2020}, Transfer Learning \cite{Wang2019}, Knowledge Distillation \cite{Li2019}, and Base+Personalization Layers \cite{Arivazhagan2019}.
	
	\item Federated mechanism: The naive algorithm of FL is FedAvg \cite{McMahan2017}, yet it often produces biased models in distributed computing. Therefore, researchers have proposed improvements to aggregation algorithms, such as FedBCD \cite{Liu2019}, SAFL \cite{Mohri2019}, FedProx \cite{Li2020}, and FedMA \cite{Wang2020}, taking into account factors like client fairness and heterogeneity. In addition to single-layer centralized aggregation, there are also approaches targeting multi-layer learning architectures, such as HierFAVG \cite{Liu2020a}, HFEL \cite{Luo2020}, FLEE \cite{Zhong2022}, and ACFL \cite{Huang2023}.
	
	\item Communication: Communication is an important concern in the field of federated learning \cite{Shah2021}, as the transmission of gradients or model parameters between clients and servers is often done wirelessly and can be highly unstable \cite{Kairouz2021}. Gradient compression \cite{Konecny2018} is a commonly used method to solve this problem.
\end{itemize}

\subsection{Federated Continual Learning}
In recent years, the issue of catastrophic forgetting in clients within the FL framework has increasingly attracted the attention of researchers \cite{yu2024overcoming}. Some scholars have proposed combining the concepts of FL and CL to develop a federated continual learning framework \cite{yang2024federated}. Yang et al. \cite{yang2024federated} systematically review the two scenarios—synchronous and asynchronous—that exist in FCL, and analyze the causes of catastrophic forgetting from both spatial and temporal dimensions. This work further clarifies the differences between FCL and traditional CL. For class-incremental problems, Dong et al. \cite{Dong2022} proposed a novel global-local forgetting compensation model, GLFC, which weakens catastrophic forgetting as much as possible from both global and local perspectives, ultimately enabling federated learning to train a globally incremental model. Qi et al. \cite{Qi2023a} proposed the FedCIL framework, which combines generative methods to use an ACGAN generator to replay synthetic data from previous distributions, thus alleviating catastrophic forgetting. Zhang et al. \cite{Zhang2023} presented TARGET to remember historical experience via knowledge distillation in class-incremental scenarios. For domain-incremental problems, Li et al. \cite{li2024sr} selected cached samples based on the importance of local samples and their relevance to the global dataset, using sample replay to overcome catastrophic forgetting. Huang et al. \cite{Huang2022} proposed a federated cross-correlation and continual learning method. To address heterogeneity issues, this method utilizes unlabeled public data for communication and constructs cross-correlation matrices to learn generalizable representations under domain shift. At the same time, for catastrophic forgetting, knowledge distillation is used in local updates to provide inter-domain and intra-domain knowledge effectively without leaking participants' privacy. In addition, some work can be applied to both class increment and domain increment scenarios. Yoon et al. \cite{Yoon2021} proposed a new federated continual learning framework called FedWeIT. This framework decomposes the local model parameters of each client into dense base parameters and sparse task-adaptive parameters to enable more efficient communication. Jiang et al. \cite{bakman2023federated} focuses on mitigating catastrophic forgetting in global models and proposes a method called Federated Orthogonal Training (FOT) to ensure orthogonal relationships between tasks. \cite{Jiang2021} proposed a federated learning architecture called Fed-Speech for the federated multi-speaker TTS system. This architecture uses progressive pruning masks to separate parameters to preserve speaker characteristics while applying selective masks to effectively reuse knowledge within tasks. Ma et al. \cite{Ma2022} presented the CFeD method based on knowledge distillation technology, which extracts old knowledge from the surrogate dataset through the construction of pseudo-labels and knowledge distillation. Additionally, some scholars have investigated client drift caused by the non-independent and identical distribution between clients during FCL \cite{venkatesha2022addressing}.

\textbf{Summary.} Our work differs from previous research in the following aspects: (1) $\rm{SacFL}$ can automatically monitor changes in data and trigger CL mechanisms without requiring extra information. (2) In addition to identifying new tasks, $\rm{SacFL}$ can also automatically discern whether a task is adversarial and activate defense mechanisms. This capability has not been considered in other works yet. (3) Different from methods like knowledge distillation, $\rm{SacFL}$ only requires storing a lightweight task-sensitive Decoder, effectively reducing storage overhead on end devices.

\section{The proposed method: SacFL}

\subsection{Motivation}\label{motivation}

\begin{figure}[!t]
	\centering
	\includegraphics[width=3.5in]{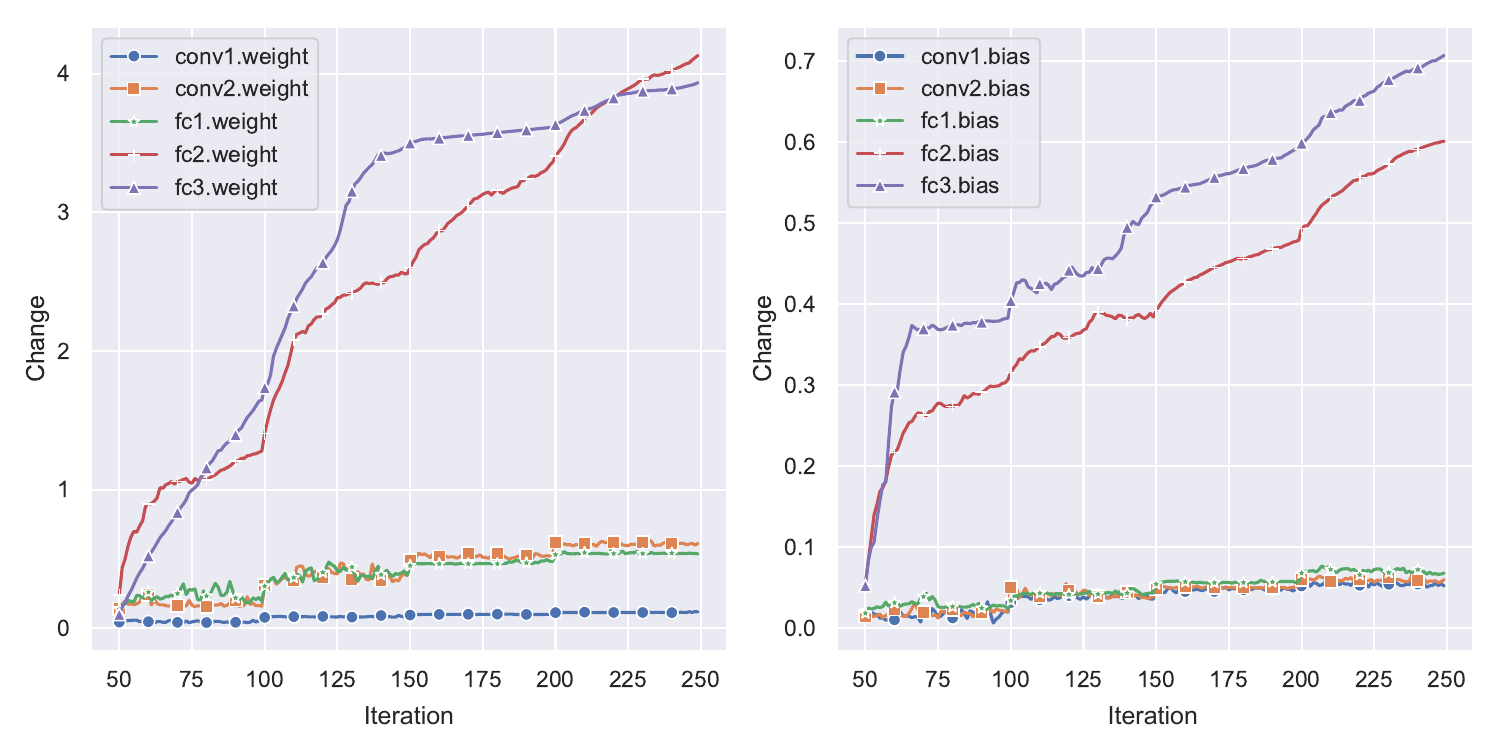}
	\caption{The changes of parameters in different model layers during the training process. It is worth noting that each task is trained for 50 iterations, and there is no need to calculate the changes in model parameters for the 0th task. Therefore, the abscissa in the figure starts from 50. The vertical axis represents the difference between specific layer parameters and the corresponding layer parameters after training the 0th task.}
	\label{layer_change}
\end{figure}

During the continual learning process, as data shifts, the last several layers of deep models (e.g., fully connected layers) change significantly, whereas the preceding layers exhibit minimal variation. Using the FashionMNIST dataset as an example, we construct a LeNet neural network comprising Convolutional Layers, Activation Layers, Max Pooling Layers, and Fully Connected Layers. Both Convolutional and Fully Connected Layers contain two types of parameters: weights and biases. In the context of continual learning, we divide the ten classes of data into five tasks, each task comprising two classes: \{0,1\}, \{2,3\}, \{4,5\}, \{6,7\}, and \{8,9\}. Each task is trained for 100 iterations, with the initialization model for each subsequent task derived from the previous one. By observing the parameter changes between consecutive tasks, we can discern the impact of task transitions on the model. In our experiment, we project multi-dimensional model parameters onto two-dimensional graphs and use the Euclidean Distance between these parameter graphs to represent changes in the model layers. The resulting curve graphs (Fig. \ref{layer_change}) illustrate the changes in weights (left) and biases (right). From these graphs, we can see that the weight and bias changes of the final fully connected layer are the most pronounced as tasks shift.

\begin{figure*}[!t]
	\centering
	\includegraphics[width=0.95\linewidth]{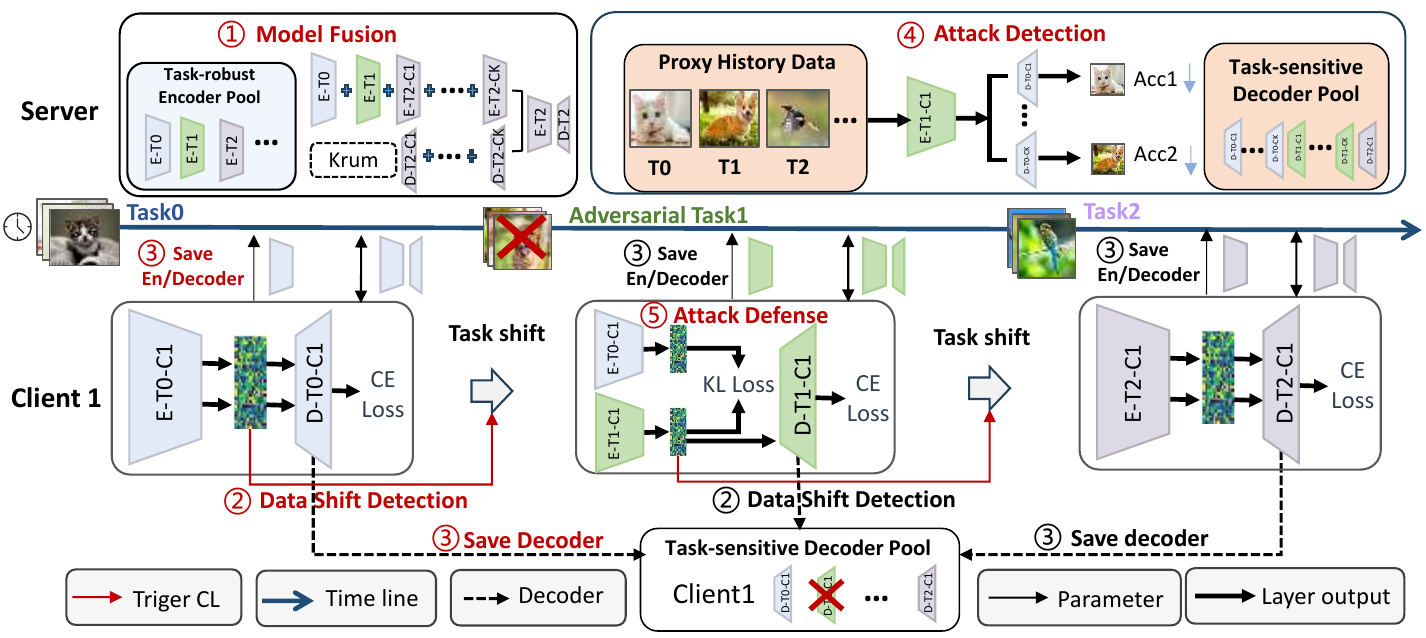}
	\caption{The framework of $\rm{SacFL}$. When no task change occurs, the client trains the Encoder and Decoder using the classical FL approach, with the exception of performing data drift detection during each iteration. If data drift is detected, the Decoder from the previous task is pushed to the local Decoder pool. At the same time, the updated Encoder and Decoder are uploaded to the server to determine whether the new task is adversarial. If an adversarial task is detected, local attack defense mechanisms and Krum aggregation are activated to mitigate the impact of the attack, continuing until the next task is identified.}
	\label{method}
\end{figure*}

\subsection{Framework and Pipeline}
\textbf{Framework.} The framework of $\rm{SacFL}$ is depicted in Fig. \ref{method}. Based on the sensitivity of model parameters to task changes, we divide the on-device model $M$ into a task-robust Encoder $E$ and a task-sensitive Decoder $D$, i.e., ${M} = {E} \circ {D}$. The parameters of the Encoder demonstrate relative stability across diverse tasks, while the Decoder shows high variability in response to task-specific dynamics. $\rm{SacFL}$ constructs an Encoder pool, a Decoder pool, and a Proxy history data pool on the server. The Encoder pool stores global Encoders for history tasks. In subsequent iterations, these history Encoders are incorporated into aggregations to release catastrophic forgetting. The Decoder pool stores Decoders for clients' history tasks, and the Proxy history data pool stores client history data collected from public sources. These two pools facilitate the monitoring of adversarial tasks. All three pools evolve as the number of tasks increases. In addition, $\rm{SacFL}$ also builds a small Decoder pool for each client to store history task Decoders, enabling rapid local access for computation.

\textbf{Pipeline.} When no task changes occur, similar to traditional FL, clients train the Encoder and Decoder using cross-entropy loss. A key difference in $\rm{SacFL}$ is that the client monitors local data drift by tracking changes in the Encoder's output after one local training epoch in each iteration. Once data drift is detected, the local task is considered to have changed, and the Decoder from the previous task is pushed to the local Decoder Pool to store history knowledge. Simultaneously, the trained Encoder and Decoder are sent to the Encoder pool and Decoder pool on the server. Server's history Decoder pool and Proxy history data are used to determine whether the new task is an adversarial task. If the new task is identified as adversarial, the attack defense strategy is implemented locally, and a robust Krum \cite{blanchard2017machine} aggregation method is applied at the server to mitigate the attack's impact until a new task is detected. It is important to note that the Decoder for adversarial tasks is not stored in the Decoder pool. When a task changes, only the Encoder is transferred between tasks, and the corresponding Decoder needs to be reinitialized at the beginning of each new task. If the user has an inference request, the relevant history Decoder is retrieved from the local Decoder pool and combined with the current Encoder to perform computation, effectively preventing catastrophic forgetting.

The proposed method offers several advantages:

(1) The Decoder typically consists of the final few layers or even a single layer. Compared to methods that store most of the history models on end devices, this approach occupies significantly less storage space, leading to substantial improvements in storage efficiency.

(2) By dividing the model into task-robust and task-sensitive layers, the task-robust layers are transferred across different tasks, ensuring the sharing of common knowledge. Meanwhile, maintaining a separate Decoder for each task preserves task independence, thereby reducing interference between tasks.

(3) The design of data drift and adversarial task detection methods enables the timely detection of task changes and self-adaptive defense against adversarial attacks. These methods enhance the client's self-adaptive continual learning.

\subsection{Training Process}\label{Training Process}
Assuming there are $K$ clients, \textit{i.e.}, end devices, in the federated learning framework. Each client faces $T$ continual learning tasks, which can be represented as $\left\{ {0, \cdots, t, \cdots, T} \right\}$ with $I_t$ federated iterations for task $t$. The total number of iterations is $\mathbb{I} = \sum\nolimits_{t = 1}^T {{I_t}} $. The client models are denoted by $M$, and the set of all client models is $\left\{ {{M_1},{M_2}, \cdots ,{M_K}} \right\}$.

Generally speaking, the federated learning process can be divided into four stages: server distribution, client local training, client upload, and server aggregation. Here, we will mainly focus on local training and server aggregation. During the client local training stage, the number of local epochs is $N$ in each round. When client $k$ faces task $t$, the trained model $M_k^t$ is obtained. Assuming the learning rate of the client is $\eta$, the client's local training process can be represented as follows:

\begin{equation}
	\label{eq1}
	M_k^t(i,n) = M_k^t(i,n - 1) - \eta \nabla F_k^t(M_k^t(i,n - 1)),{\rm{        }}n = 1, \cdots ,N,
\end{equation}
where $M_k^t(i,n)$ represents the model obtained from the $n$-th local epoch of client $k$ in the $i$-th iteration of task $t$. $F_k^t(M_k^t(i,n - 1))$ denotes the loss function of the model $M_k^t(i,n-1)$ when client $k$ faces task $t$. Before the client starts local training, the model parameters obtained from the server side are $M_k^t(i,0)$, and they can be represented as:
\begin{equation}
	\label{eq2}
	M_k^t(i,0) = E_k^t(i,0) \circ D_k^t(i,0),
\end{equation}
where,
\begin{equation}
	\label{eq3}
    E_k^t(i,0) = \frac{\sum\nolimits_{j = 0}^{t - 1} {E_{.}^j({I_j},N) + E_{.}^t(i,0)} }{t + 1},
\end{equation}
\begin{equation}
	\label{eq6}
	E_{.}^t(i, 0)=\sum_{k=1}^K \frac{D S_k^t}{\sum_{k=1}^K DS_k^t} E_k^t(i-1, N),
\end{equation}

\begin{equation}
	\label{eq4}
	D_k^t(i, 0)=D_{.}^t(i, 0)=\sum_{k=1}^K \frac{D S_k^t}{\sum_{k=1}^K DS_k^t} D_k^t(i-1, N).
\end{equation}

$E_k^t(i, 0)$ undergoes a two-stage fusion. The first stage is spatial fusion (refer to Eq. (\ref{eq6})). In Eq. (\ref{eq6}), $E_{.}^t(i, 0)$ is the globally aggregated Encoder obtained after $i-1$ iterations at current task $t$, which is the weighted sum of $E_{k}^t(i-1, N)$. $DS_{k}^{t}$ is the data size of client $k$ during task $t$. The second stage is temporal fusion (refer to Eq. (\ref{eq3})). In Eq. (\ref{eq3}), $E_{.}^{t-1}(I_{t-1}, N)$ is the globally aggregated Encoder after $I_{t-1}$ iterations of task $t-1$ which is stored in Task-robust Pool. Note that when clients are facing the first task and there are no previous tasks, the training process of Encoder is similar to traditional federated learning steps. Eq. (\ref{eq4}) illustrates the training process of Decoders. When clients encounter a new task, they re-initialize the Decoders and then update them in a regular FL manner. After one specific task training is completed, its corresponding lightweight Decoder is stored in Task-sensitive Pools. In the above CL process, the shared knowledge contained in different tasks is inherited between generations of Encoders. Only one Encoder needs to be stored in the clients to inherit the common knowledge of historical tasks, while a memory-efficient branch is dedicated to storing task-sensitive knowledge. This approach significantly reduces end devices' storage requirements and promotes long-term CL.

It is worth noting that in the process of CL, the structure of $E_{\cdot}^{t}(\cdot,\cdot)$ and $E_{\cdot}^{t-1}(\cdot, \cdot)$ remains the same, but the structure of $D_{\cdot}^{t}(\cdot, \cdot)$ and $D_{\cdot}^{t-1}(\cdot, \cdot)$ does not always remain consistent. For example, in class-incremental tasks, when the model encounters more classes, the branch structure of the model will automatically expand to adapt to the new task, resulting in a significant change in $D_{\cdot}^{t}(\cdot, \cdot)$ structure.

\subsection{Data Drift Detection}\label{Data Drift Detection}
The traditional method for data drift detection relies on data comparison or performance observation. However, these methods require a considerable amount of memory to store historical data or labeled data, which is not friendly for resource-constrained end devices. Meanwhile, in the context of $\rm{SacFL}$, the method proposed in Section \ref{Training Process} is a model-based CL technique that does not store historical data; the client only retains data for the current task. Therefore, inspired by contrastive learning \cite{Li2021}, we propose a memory-efficient and label-free data drift detection method. Data drift can be detected by comparing the Encoders' outputs before and after local learning on clients. The specific approach is as follows: after the server distributes the aggregated Encoder to clients, each client performs one round of local training using its local data. To measure the difference between the Encoders before and after local training, a certain number of current task data are picked and input into the above two models. If the change value exceeds a certain threshold, it indicates significant differences in the features extracted by the two models from the same data. We can then conclude that substantial model changes and data drift have occurred on the client.

In this process, it is important to note that we use the difference between the output features of Encoders to detect data shift. Since the Encoder is less sensitive to data alterations compared to the Decoder, data drift is only identified when the Encoder's output undergoes substantial changes, preventing misjudgments and improving the accuracy of data shift detection. Furthermore, experiments reveal that compared to commonly used Euclidean distance and Cosine distance, the Manhattan distance is more sensitive to variations in the Encoder's output (as shown in Fig. (\ref{Diff})). From Fig. (\ref{Diff}), we can see that the variations of Euclidean distance and Cosine distance are less than Manhattan distance when the task shifts. However, when the task does not change, the value of Manhattan distance remains nearly unchanged. Therefore, we employ the Manhattan distance to detect data drift. The calculation formula is as follows:

\begin{equation}
	\label{eq5}
\operatorname{Diff}=\operatorname{Manhattan}\left(E_k^t(i, 1)\left(D A_k^t\right), E_k^t(i, 0)\left(D A_k^t\right)\right),
\end{equation}
where $E_k^t(i, 1)$ represents the Encoder parameters of the $k$-th client after locally training one epoch during $i$-th federated iteration when facing task $t$. $D A_k^t$ refers to the data for task $t$ of client $k$, and $E_k^t(i, 1)\left(D A_k^t\right)$ represents the data features extracted by inputting $D A_k^t$ into the Encoder $E_k^t(i, 1)$. $E_k^t(i, 0)$ is the received Encoder of client $k$ at the beginning of iteration $i$ when facing task $t$. Similarly, $E_k^t(i, 0)\left(D A_k^t\right)$ denotes the extracted feature of $E_k^t(i, 0)$ by inputting $D A_k^t$. This method is effective not only when the new data is benign, but also demonstrates its efficacy in adversarial tasks, as validated in Section \ref{Data Drift Detection exp}. It can accurately identify adversarial data as a new task. Through the data detection mechanism, end devices can automatically detect data changes and trigger CL, greatly enhancing the clients' self-adaptive capabilities.

\subsection{Adversarial Attack Defense}\label{Adversarial Attack Defense}
In the process of CL, new tasks may involve adversarial examples aimed at attacking the model. Therefore, when a new task arises, it should be assessed first. Only when the new task's samples are benign should the CL mechanism be activated. If the new task consists of adversarial data, appropriate defense measures are needed to mitigate the impact on historical knowledge. Accordingly, we propose methods for adversarial task detection and adversarial attack defense.

\textbf{Adversarial Task Detection.} In $\rm{SacFL}$, we construct a Decoder pool for history tasks and a Proxy history data pool on the server for adversarial task monitoring. Suppose client $z$ detects a switch from task $j-1$ to task $j$ using a data drift detection mechanism. The updated Encoder $E_z^j(0,1)$ is uploaded to the server. Then, the updated Encoder is combined with the Decoders from the Decoder pool ${P_D} = \{D_k^t(I_t, N) |k \in [0, K],t \in [0,j)\}$, respectively. After that, the corresponding proxy history data is fed into the model, generating the following outputs:
\begin{equation}
\label{eq7}
Opt_k^t = ( {E_z^j(0,1)^\circ D_k^t ( {{I_t},N})} )( {x_k^t,y_k^t}),{\rm{    }}\forall k \in  [0,K] ,\forall t \in [0,j-1].
\end{equation}

Based on the above outputs, the accuracy on all clients $k \in [0, K]$ and the corresponding historical tasks $t \in [0,j-1]$ is obtained, from which the performance degradation rate of the historical task caused by $E_z^j(0,1)$ is calculated:
\begin{equation}
\label{eq8}
Degrade_z^j = \frac{1}{K}\sum\nolimits_{k = 1}^K {\left( {\frac{1}{j}\sum\nolimits_{t = 0}^{j-1} {\frac{{Acc_k^t - \widetilde {Acc}_k^t}}{{Acc_k^t}}} } \right)},
\end{equation}
$Acc_k^t$ represents the original accuracy of client $k$ on task $t$. When $Degrade_z^j$ exceeds a certain threshold, we consider the task to be adversarial. This is because, the Encoder's parameter changes a little, and the historical Decoder is used for testing, which, in principle, should not cause a significant degradation in performance on historical tasks. If a significant performance drop in the historical task still occurs, it indicates that the new task is adversarial, directly leading to a substantial change in the Encoder's parameters.

\textbf{Adversarial Attack Defense.} To effectively defend against the above-mentioned attacks, we constrain the changes in the Encoder during the training of adversarial tasks. Suppose client $z$ detects task $j$ as an adversarial task, while task $j-1$ is a benign task. In this case, the KL divergence between the output of $E_{.}^{j - 1}({{\rm{I}}_{j - 1}}, E)$ and $E_z^j({\rm{i}},e)$ is computed. By minimizing this value, the degree of performance degradation can be reduced. The formula for this calculation is as follows:
\begin{equation}
\label{eq9}
{{\cal F}_{ebd}} = KL(E_{.}^{j - 1}({{\rm{I}}_{j - 1}},E)(x_z^j,y_z^j),E_z^j({\rm{i}},e)(x_z^j,y_z^j)).
\end{equation}

At the same time, the cross-entropy loss should also be considered:
\begin{equation}
    \label{eq10}
    {{\cal F}_{ce}} = CE(y_z^j,M_z^j({\rm{i}},e)(x_z^j,y_z^j)),
\end{equation}

Finally, we get the local training loss:
\begin{equation}
    \label{eq11}
    {\cal F}_k^t = \alpha {{\cal F}_{ebd}} + (1 - \alpha ){{\cal F}_{ce}}.
\end{equation}

In addition, we also employ a more robust aggregation method, Krum \cite{blanchard2017machine}, on the server to defend against adversarial attacks.

\subsection{Algorithm}
To elucidate the method described above, we provide an algorithmic explanation in Algorithm. \ref{alg: SacFL}. The algorithm's inputs include the number of clients $K$, the total number of federated learning iterations $\mathbb{I}$, the number of local training rounds $N$, the data for each client $(x_k^t,y_k^t)$, the learning rate $\eta$, the Encoder pool $P_E$ and Decoder pool $P_D$ on the server, and Proxy history data pool $P_{pd}$. The final output is the global Encoder and task-sensitive Decoders. 

Initially, the server initializes the task ID and $M_{\cdot}^t$ (Algorithm \ref{alg: SacFL}. Lines 1-2), and then separates the model into Encoder and Decoder based on the layer changes with task shifts (Algorithm \ref{alg: SacFL}. Line 3). The Encoder shows low sensitivity to task variations, while the Decoder is highly sensitive. Subsequently, the initialized Encoder and Decoder are distributed to the clients (Algorithm \ref{alg: SacFL}. Line 5). Upon receiving the model, each client performs $N$ rounds of local training. When the federated iteration count is greater than 1, each client checks for data shift (Algorithm \ref{alg: SacFL}, Line 12) after one epoch of local training. If a task change is detected, the $E_{k}^t(i,1)$ remains unchanged, but the $D_{k}^{t}(i, 1)$ is reinitialized, and the Task-sensitive Decoder Pool is updated (Line 15-16). After that, clients will upload $E_{k}^t(i,1)$ to the server for further detection to determine whether it is an adversarial task (Lines 17-19). If it is identified as an adversarial task, local defensive training will be conducted using Eq. (\ref{eq11}) (Lines 20-21). After $N$ rounds of local training, the clients upload their Encoder $E_{k}^t(I, N)$ and Decoder $D_{k}^{t}(i, N)$ to the server (Line 22). At this stage, if data drift occurs on the client side, it is necessary to update both the iteration $i$ of the current task and the task ID $t$ (Lines 23-25). Then, the server selects different strategies to aggregate all Encoders and Decoders based on whether the clients are under attack. Finally, $E_{.}^t(i + 1,0)$ and $D_{.}^t(i + 1,0)$ are obtained (Lines 27-30) and the Encoder pool is updated (Line 31).

\begin{algorithm}[!t]
	\SetAlgoVlined
	\small
	\KwIn{Clients' number $K$, learning rate $\eta$, federated round $\mathbb{I}$, clients' model $M_{\cdot}^0$ (composed by Encoder $E_{\cdot}^0$ and Decoder $D_{\cdot}^0$), local epoch $N$, client $k$'s data $(x_k^t,y_k^t)$, Decoder pool on the server $P_D$, Encoder pool $P_E$, Proxy history data pool $P_{pd}$}
	\KwOut{Task-robust Encoder, Task-sensitive Decoder Pool}
    Initialize task ID $t=0$; \\
	Initialize global model $M_{\cdot}^t$ on the server;\\
	Decompose model $M_{\cdot}^t$ into Encoder $E_{\cdot}^t$ and Decoder $D_{\cdot}^t$;\\
	\For{federated round $i=1,\cdots,\mathbb{I}$}{Server distribute $E_{\cdot}^t(i)$, $D_{\cdot}^t(i)$ to clients;\\
    // Local Training\\
		\For{client $k=1,\cdots,K$}{\For{local epoch $n=1,\cdots,N$}{$M_k^t(i,n) \leftarrow M_k^t(i,n - 1) - \eta F_k^t((x_k^t,y_k^t),$\\$M_k^t(i,n - 1))$;\\
				\If{$i > 1$ and $n=1$}{$SHIFT \gets DataDetection((x_k^t,y_k^t),$\\$E_{k}^{t}(i,0),E_{k}^{t}(i,1))$;\\
					\If{$SHIFT = True$}{
                    Initialize $D_{k}^{t}(i, 1)$ as $D_{k}^{t+1}(0, 0)$;\\
                    Push $D_{k}^{t}(i-1, N)$ to Decoder Pools on the client $k$ and the server;\\
                    Push $E_{k}^{t}(i, 1)$ to the server for attack detection;\\
                    $ATTACK \gets AttackDetection(E_{k}^{t}$\\$(i,1), P_{ph}, P_D)$;\\
                    \If{$ATTACK = True$}{
                    Local updating using Eq. (\ref{eq11})}
                    }}}
			Upload $E_{k}^{t}(i,N)$ and $D_{k}^{t}(i,N)$ to the server;}
		$i$$\gets$$i+1$;\\
        \If{$SHIFT = True$}{$t$$\gets$$t+1$, $i$$\gets$$1$}
		// Server Aggregation\\
        \If{$ATTACK = True$}{
    $E_{\cdot}^t(i + 1,0),D_{\cdot}^t(i + 1,0)\gets$Aggregation using $P_E$ based on Krum;
}
\Else{
$E_{\cdot}^t(i + 1,0),D_{\cdot}^t(i + 1,0)\gets$Aggregation using $P_E$ based on Eq. (\ref{eq3}) and Eq. (\ref{eq4});
}
Update $P_E$ with $E_{\cdot}^t(i + 1)$.}    
	\small
	\caption{$\rm{SacFL}$}
	\label{alg: SacFL}
\end{algorithm}

\section{Theoretical Analysis}\label{Theoretical analysis}
In $\rm{SacFL}$, when there are no changes in the client, the convergence analysis is similar to that of FedAvg. However, differences arise when clients autonomously switch to different tasks. First, during the aggregation process, it is necessary not only to aggregate the current client models but also to integrate historical task models. Second, the autonomy of task switching among different clients leads to noticeable differences in distributions between clients. To demonstrate the convergence of $\rm{SacFL}$ in the context of CL, it is essential to establish the convergence of each sub-task in this scenario. Therefore, we begin with an analysis of sub-task $t$.

\textbf{Aiming at the first difference}, we regard all historical models as client models that do not participate in training but are solely involved in the aggregation process. It can be derived by following formulas:

\begin{equation}
\begin{aligned}
    E_k^t(i,0) &= \frac{{\sum\nolimits_{j=0}^{t - 1} {E_{.}^j({I_j},N) + E_{.}^t(i,0) }}}{{t + 1}} \\
    &= \sum\nolimits_{j = 0}^{t - 1} {\frac{1}{{t + 1}}E_{.}^j({I_j},N)} \\
    &+ \sum\limits_{k = 1}^K {\frac{{DS_k^t}}{{(t + 1)\sum\nolimits_{k = 1}^K {DS_k^t} }}} E_k^t(i - 1,N),
\end{aligned}
\end{equation}
where the aggregated weight of historical models is $\frac{1}{{t + 1}}$.

\textbf{Aiming at the second difference}, we have following assumptions and definition:

\begin{assumption}
	For task $t$, (1) all clients participate in training; (2) $F_k^t$ is $Z^t$-smooth and $\gamma^t$-convex; (3) the expected variance of client $k$'s stochastic gradients is bounded by $(\beta_k^t)^2$; (4) the expected value of the square of client $k$'s stochastic gradients' norm is bounded by $(\rho^t)^2$.
\end{assumption}

\begin{definition}
	Define $\phi^t$ as the heterogeneity degree of data shift, which is calculated as follows:
    \begin{equation}
          \phi^t = {\tilde F^t} - \sum\nolimits_{j = 1}^K \frac{D{S_k^t}}{\sum\nolimits_{k = 1}^K D{S_k^t}} \tilde F_k^t,
    \end{equation}
	where $\tilde F^t$ and $\tilde F_k^t$ are the minimum of $F^t$ and $F_k^t$, respectively.
\end{definition}

\textbf{If we want to prove the global model on task $t$ is convergent, then the following inequation should be satisfied:}
\begin{equation}
\label{prove obj}
[{F^t}(M_{.}^t({\rm{i}}))] - {\tilde F^t} \le an upper bound \mathbb{B},
\end{equation}
where $I$ is the iteration number of task $t$, $\tilde F^t$ is the optimal loss value. When $\mathbb{B}$ decreases as the number of iterations $i$ increases, it indicates that the global model is progressively approaching the optimal model for task $t$.

According to Assumption (2), $Z^t$-smooth function $F_k^t$ possesses the following properties:
\begin{equation}
\begin{aligned}
    F_k^t\left( {M_.^t(i)} \right) \le F_k^t(\tilde M_.^t) &+ {(M_.^t(i) - \tilde M_.^t)^T}\nabla F_k^t(\tilde M_.^t)\\ 
    &+ \frac{{{Z^t}}}{2}{\left\| {M_ \cdot ^t(i) - \tilde M_ \cdot ^t} \right\|^2},
\end{aligned}
\end{equation}
where $\tilde M_.^t$ is the parameter that minimizes the loss value, and its gradient is $\nabla F_k^t(\tilde M_.^t) = 0$. Therefore, the above equation can be further transformed into:

\begin{equation}
\label{theoretical1}
\mathbb{E}\left[ {F_k^t\left( {M_.^t(i)} \right)} \right] - F_k^t(\tilde M_.^t) \le \frac{{{Z^t}}}{2}{\mathbb{E}\left\| {M_{.}^t(i) - \tilde M_{.}^t} \right\|^2}.
\end{equation}

In the above formula, $\frac{{{Z^t}}}{2}{\mathbb{E}\left\| {M_{.}^t(i) - \tilde M_{.}^t} \right\|^2}$ is $\mathbb{B}$ in Eq. (\ref{prove obj}). Since $Z^t$ is a constant, we only need to prove that ${\mathbb{E}\left\| {M_{.}^t(i) - \tilde M_{.}^t} \right\|^2}$ decreases with the number of iterations $i$ increases, in order to achieve global convergence. Based on Assumptions (3)-(5) and definitions, combined with Lemma 1-3 from reference \cite{Li2019a}, it can be derived that:
\begin{equation}
\label{theoretical2}
{\mathbb{E}\left\| {M_{.}^t(i) - \tilde M_{.}^t} \right\|^2} \le \frac{{{\lambda ^t}}}{{{\zeta ^t} + i}},
\end{equation}
where ${\lambda ^t} = \max \left\{ {\frac{{{\mu ^2}{G^t}}}{{\mu {\gamma ^t} - 1}},({\zeta ^t} + 1){{\mathbb{E}\left\| {M_{.}^t(1) - \tilde M_{.}^t} \right\|}^2}} \right\}$, $\mu$ is some value larger than $\frac{1}{\gamma ^t}$, ${\zeta ^t} = \max \{ \frac{{8{Z^t}}}{{{\gamma ^t}}},N\}$, and ${G^t} = \sum\limits_{k = 1}^K {{{\left( {\frac{{DS_k^t}}{{\sum\nolimits_{k = 1}^K {DS_k^t} }}} \right)}^2}} {\left( {\beta _k^t} \right)^2} + 6{Z^t}{\phi ^t} + 8{(N - 1)^2}{\left( {{\rho ^t}} \right)^2}$.

When $\mu  = \frac{2}{{{\gamma ^t}}}$, then,
\begin{equation}
\label{theoretical3}
\begin{aligned}
    {\lambda ^t} &= \max \{ \frac{{{\mu ^2}{G^t}}}{{\mu {\gamma ^t} - 1}},({\zeta ^t} + 1){\mathbb{E}\left\| {M_ \cdot ^t(1) - \tilde M_ \cdot ^t} \right\|^2}\} \\
 &\le \frac{{{\mu ^2}{G^t}}}{{\mu {\gamma ^t} - 1}} + ({\zeta ^t} + 1){\mathbb{E}\left\| {M_ \cdot ^t(1) - \tilde M_ \cdot ^t} \right\|^2}\\
 &= \frac{{4{G^t}}}{{{{\left( {{\gamma ^t}} \right)}^2}}} + ({\zeta ^t} + 1){\mathbb{E}\left\| {M_ \cdot ^t(1) - \tilde M_ \cdot ^t} \right\|^2}.
\end{aligned}
\end{equation}

Combining Eq. (\ref{theoretical1}), Eq. (\ref{theoretical2}), and Eq. (\ref{theoretical3}), we can get

\begin{equation}
    \begin{array}{l}
\mathbb{E}[F_k^t(M_.^t(i))] - F_k^t(\tilde M_.^t)\\
 \le \frac{{{Z^t}}}{{{\zeta ^t} + i}}[\frac{{2{G^t}}}{{{{\left( {{\gamma ^t}} \right)}^2}}} + \frac{{({\zeta ^t} + 1){{\mathbb{E}\left\| {M_ \cdot ^t(1) - \tilde M_ \cdot ^t} \right\|}^2}}}{2}].
\end{array}
\end{equation}

From the above equation, it can be observed that for a single task $t$, as the number of iterations $i$ increases, the loss values of the aggregated global model across various clients gradually decrease and approach the minimum value. \textbf{Therefore, it can be concluded that $\rm{SacFL}$ converges for each task $t$ within the framework of CL, leading to overall convergence in the CL process.}

Furthermore, ${G^t}=\sum\limits_{k = 1}^K {{{\left( {\frac{{DS_k^t}}{{\sum\nolimits_{k = 1}^K {DS_k^t} }}} \right)}^2}} {\left( {\beta _k^t} \right)^2} + 6{Z^t}{\phi ^t} + 8{(N - 1)^2}{\left( {{\rho ^t}} \right)^2}$, we obtain:
\begin{equation}
\begin{aligned}
&\mathbb{E}[F_k^t(M_.^t(i))] - F_k^t(\tilde M_.^t)= \frac{{12{Z^{2t}}}}{{\left( {{\zeta ^t} + i} \right){{\left( {{\gamma ^t}} \right)}^2}}}{\phi ^t}\\ 
&+ \frac{{2{Z^t}\left( {\sum\limits_{k = 1}^K {{{\left( {\frac{{DS_k^t}}{{\sum\nolimits_{k = 1}^K {DS_k^t} }}} \right)}^2}} {{\left( {\beta _k^t} \right)}^2} + 8{{(N - 1)}^2}{{\left( {{\rho ^t}} \right)}^2}} \right)}}{{{{\left( {{\gamma ^t}} \right)}^2}({\zeta ^t} + i)}}\\
&+\frac{{({\zeta ^t} + 1){Z^t}{{\mathbb{E}\left\| {M_ \cdot ^t(1) - \tilde M_ \cdot ^t} \right\|}^2}}}{{2({\zeta ^t} + i)}}.\\
\end{aligned}
\end{equation}

From the above equation, it can be seen that as the CL progresses, the tasks autonomously vary among different clients, leading to increased $\phi ^t$. This enhancement in heterogeneity results in a greater number of iterations required for convergence, thereby slowing down the convergence rate.

\section{Experimental Verification}\label{experiments}
In the experimental section, we mainly focus on answering the following questions:

\textbf{\textit{Q1}}: Under the federated learning framework, is the $\rm{SacFL}$ effective compared to mainstream continual learning methods when the client's task changes occur infrequently?

\textbf{\textit{Q2}}: In scenarios where the client's task undergoes continuous changes, does $\rm{SacFL}$ maintain its advantages?

\textbf{\textit{Q3}}: Apart from class-incremental learning, does $\rm{SacFL}$ retain its effectiveness when the clients' data experiences domain-incremental changes?

\textbf{\textit{Q4}}: When a new task involves adversarial data, how can clients defend against them?

\textbf{\textit{Q5}}: Can $\rm{SacFL}$ reduce resource consumption on end devices compared to other continual learning methods?

\textbf{\textit{Q6}}: What is the impact of the data drift detection mechanism on model performance?

\textbf{\textit{Q7}}: Does $\rm{SacFL}$ still perform well in the demo system from the real world?

The answers to the above questions correspond to Section \ref{Simple Continual Learning}, \ref{Sequential Continual Learning}, \ref{Domain Continual Learning}, \ref{adversarial CL}, \ref{Resource Analysis}, \ref{Ablation Studies} and \ref{Real Machine Experiments} respectively. Our code is available at: https://github.com/Zhong-Zhengyi/SacFL-Code.

\subsection{Experimental Settings}
\subsubsection{\textbf{Framework}}
To answer the above-mentioned questions, we design a federated learning framework consisting of fifty clients and one server. This framework is tailored for the cross-device scenario in federated learning, wherein a subset of clients participates in each iteration round. To minimize the consumption of client storage resources during CL, we utilize the last layer of the model as the Decoder, while all preceding layers serve as the Encoder in our experiments.

\subsubsection{\textbf{Datasets}}
The experimental image datasets encompass FashionMNIST, Cifar10 \cite{Krizhevsky2009}, and Cifar100 \cite{Krizhevsky2009}, featuring 10, 10, and 100 classes respectively. Additionally, the text dataset employed is THUCNews \cite{Li2006}, comprising 14 categories of Chinese news data collected from Sina News RSS between 2005 and 2011. To cover cases where the number of classes between tasks is equal (task num=5) and unequal (task num=3), we select 10 classes and randomly sample 5000 news from each class. These classes include lottery, stock, education, furnishment, technology, fashion, sports, game, social, and entertainment. Among these, 4000 are designated for training, while the remaining 1000 are reserved for testing.

\subsubsection{\textbf{CL Settings}}
\begin{figure}[!ht]
	\centering
	\includegraphics[width=3.5in]{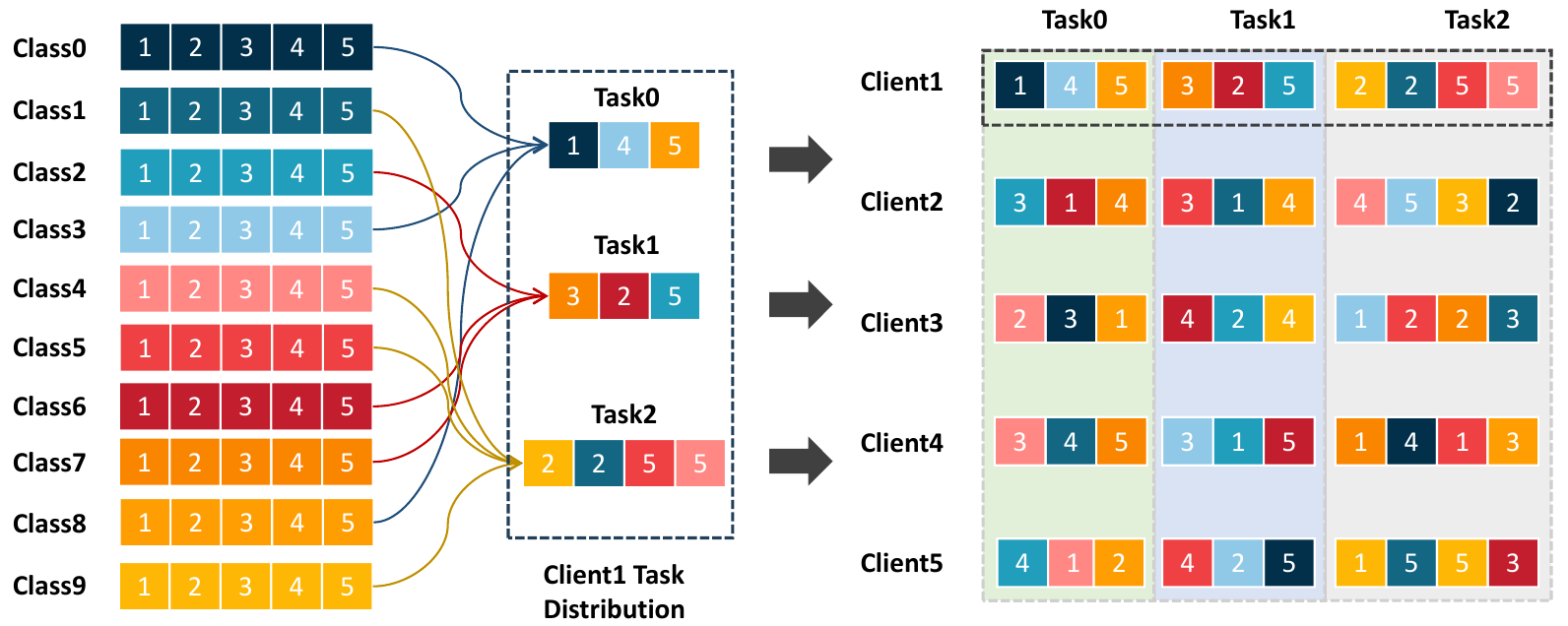}
	\caption{Class-incremental data setting. Taking 5 clients as an example, the data of each class is divided into 5 parts. Through random selection, Client 1 extracts the 1st, 4th, and 5th parts from the data labeled 0, 3, and 8 as the data for Task 0. Subsequent tasks are generated in the same manner.}
	\label{data_distri}
\end{figure}

To address the class-incremental problem, referring to the experimental setup of Qi et al. \cite{Qi2023}, we split the data classes into $T$ parts, corresponding to the total number of tasks. For example, if there are 3 tasks and 10 classes in total, each task comprises 3, 3, and 4 classes, respectively. More specifically, if the label set for the first task of client 1 is \{0, 3, 8\}, and for the second task, it is {7, 6, 2}, thus, the third task comprises classes 9, 1, 5, and 4; in contrast, if there are 5 tasks, each task comprises 2 classes. It should be noted that in real-life scenarios, data classes across different clients may intersect. Therefore, to better simulate real-world situations, we randomly sample a specific number of data classes for each client in one task. Meanwhile, an equal amount of data from the same class is randomly distributed among the clients to prevent duplication. The detailed process is illustrated in Fig. \ref{data_distri}. This approach ensures coverage of two data distribution scenarios between clients: iid and non-iid. In addition, to address the domain-incremental problem, we opt to introduce Gaussian noise and multiplicative noise to simulate domain-incremental scenarios. Similar to \cite{Ma2022} and \cite{Yoon2021}, we evaluate the effectiveness of CL by measuring the model's average testing accuracy on the current task and historical tasks. A lower accuracy indicates more severe catastrophic forgetting.

\subsubsection{\textbf{Baselines}}
The benchmarks consist of two categories: continual-based methods and traditional methods. The continual-based methods include CFeD \cite{Ma2022}, LwF-Fed \cite{Li2017}, EWC-Fed \cite{Kirkpatrick2017}, MultiHead-Fed \cite{Ma2022}, FCIL \cite{Dong2022} and FedWeIT \cite{Yoon2021}. The traditional federated methods mainly include two classic algorithms in the federated learning field: \textbf{FedAvg} \cite{McMahan2017} and \textbf{FedProx} \cite{Li2020}.
\subsubsection{\textbf{Hyper-parameters}}
We selected Adam as the optimizer with a batch size of 32, a learning rate of 0.05 for FashionMNIST and THUNews, a learning rate of 0.01 for Cifar10/100,
and a local training epoch number of 5. For the number of iterations of a single task, FahionMNIST and Cifar10 are 100, THUNews is 50, and Cifar100 is 50 or 100.

\subsection{Data Drift \& Adversarial Task Detection}\label{Data Drift Detection exp}
\begin{figure*}[htbp]
	\centering
	\subfloat[Fa-MNIST,task num=3]{\includegraphics[width=0.32\linewidth]{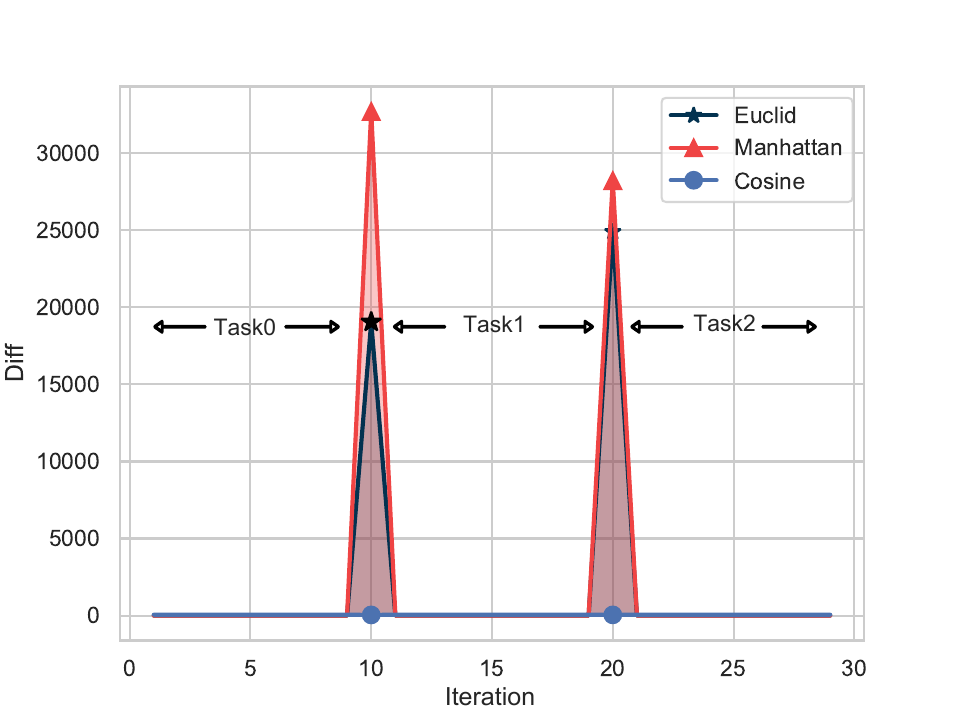}%
		\label{Diff_LeNet_FashionMNIST_3}
	}
	\hfil
	\subfloat[Cifar10,task num=3]{\includegraphics[width=0.32\linewidth]{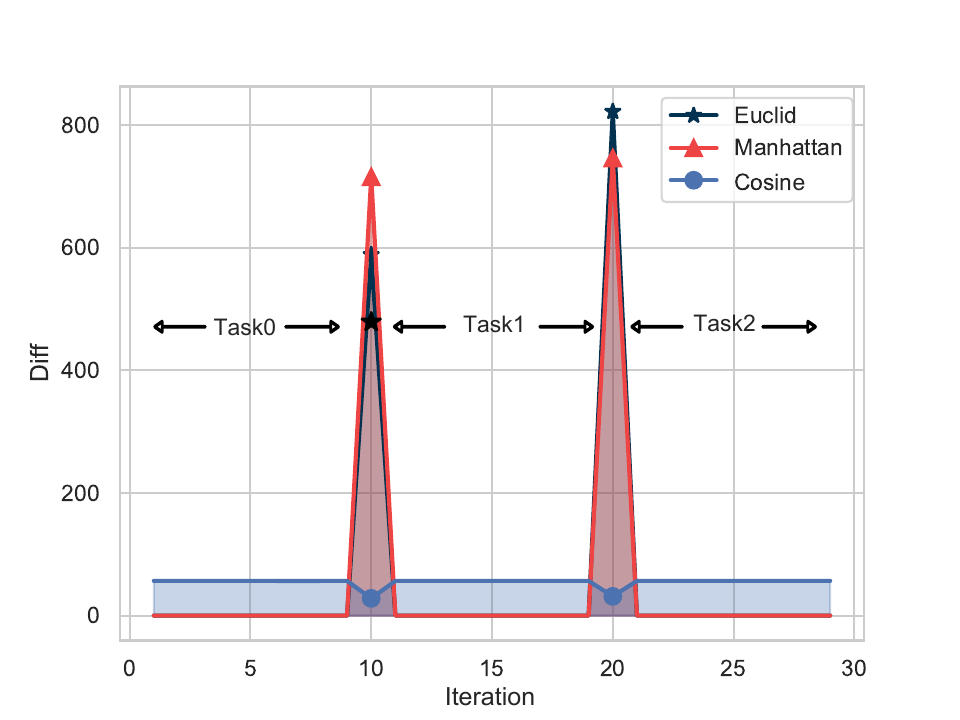}%
		\label{Diff_Cifar10_CNN_3}
	}
	\hfil
	\subfloat[THUCNews,task num=3]{\includegraphics[width=0.32\linewidth]{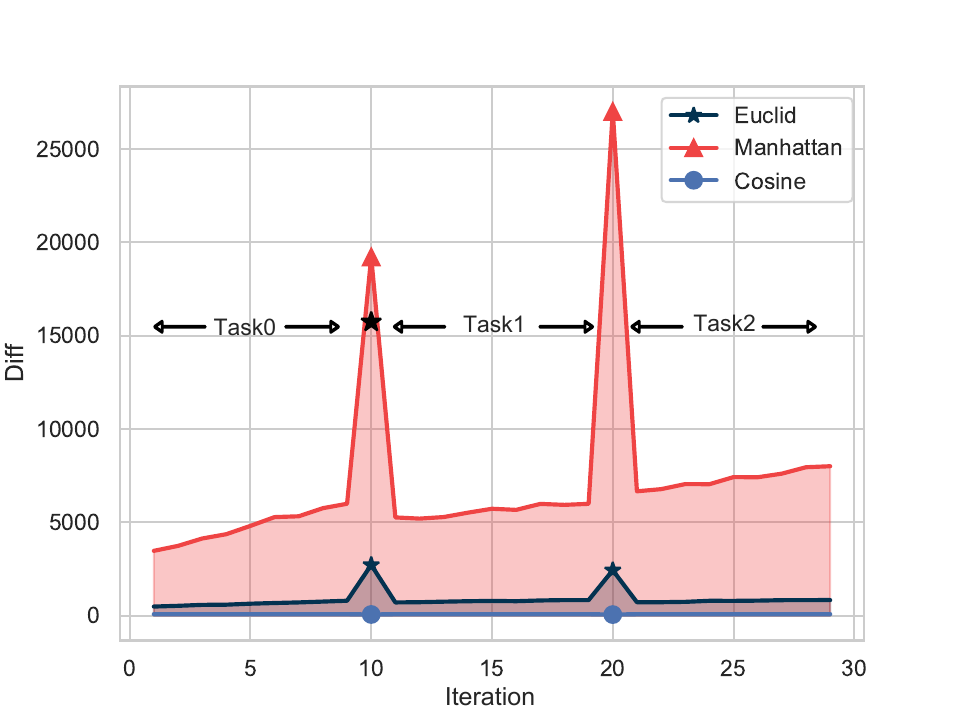}%
		\label{Diff_TextCNN_3}
	}
	\caption{The change of the feature values extracted by the Encoder.}
	\label{Diff}
\end{figure*}

\textbf{Data Drift Detection.} In real-world scenarios, data changes frequently happen without clear indicators. Hence, it's necessary to design an appropriate data drift detection mechanism to identify these changes and trigger the continual learning process. This section focuses on investigating threshold configurations for activating the continual learning mechanism. The goal is to equip $\rm{SacFL}$ with the capability to accurately detect dataset shifts, thus facilitating subsequent continual learning tasks. Fig. \ref{Diff} illustrates the changes in Encoder features for the FashionMNIST, Cifar10, and THUCNews datasets under the 3-task scenario. In our experiment, we conducted 10 federated rounds for each task. From the figures, it can be observed that the Encoder's extracted features exhibit sharp fluctuations during task transitions, indicating significant changes. Specifically, with a total of 3 tasks, the Mahattan values of the Encoder features for FashionMNIST increase from nearly 0 to over 20000, for Cifar10 from almost 0 to over 600, and for THUCNews from approximately 5000 to over 15000. Consequently, we set the threshold at 20000 for FashionMNIST, 600 for Cifar10, and 15000 for THUCNews. Following the first local training epoch at clients, if the change values of Encoders' extracted features surpass the specified thresholds, we identify data shifts.

\begin{figure}[!ht]
	\centering
	\includegraphics[width=1.0\linewidth]{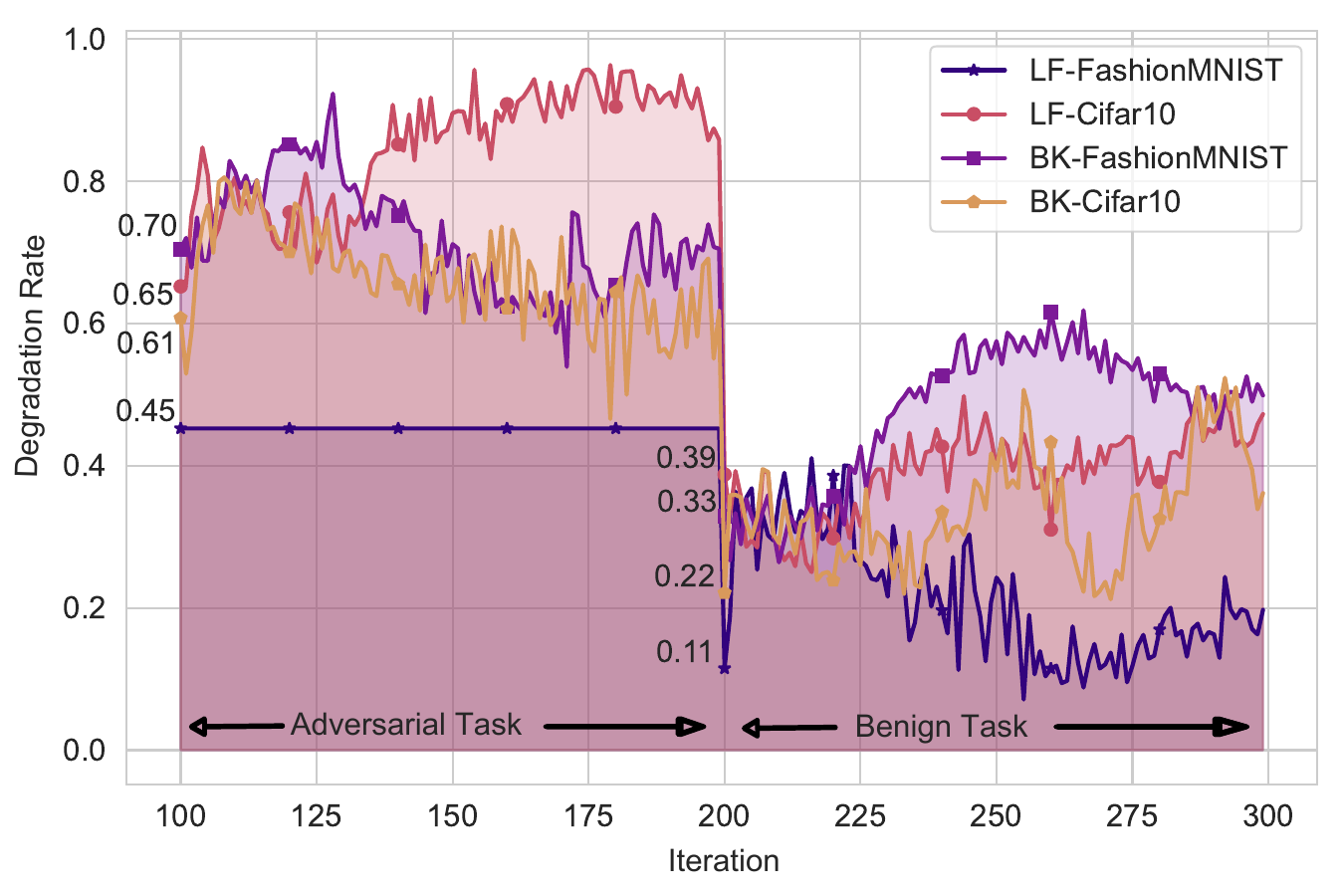}
	\caption{Average degradation rate of historical tasks under adversarial attacks.}
	\label{attack_detection}
\end{figure}

\textbf{Adversarial Task Detection.} During the CL process, when new samples are adversarial, they significantly degrade the performance on history tasks compared to general catastrophic forgetting. Therefore, after detecting data drift, it is necessary to further confirm whether the data is adversarial. This section validates the adversarial task detection mechanism using the FashionMNIST and Cifar10 datasets, under non-targeted attacks (label flipping) and targeted attacks (backdoor attacks). The number of tasks is 5, with task 1 being the adversarial data. By observing the decline of historical knowledge, we can identify the adversarial task. As shown in Fig. \ref{attack_detection}, the vertical axis represents the average degradation rate of the model's performance on the proxy historical data. At the beginning of label flipping (Iteration=100), the degradation rates for Cifar10 and FashionMNIST are 65\% and 45\%, respectively, which are higher than the benign tasks (Iteration=200) with initial degradation rates of 39\% and 11\%. When the attack method is a backdoor attack (Iteration=100), the initial degradation rates for Cifar10 and FashionMNIST are 61\% and 70\%, respectively, again exceeding that for benign tasks (Iteration=200), which are 22\% and 33\%. Overall, regardless of the type of attack, the degradation rates at the beginning of attacks (Iteration=100) are above 40\%, while these initial degradation rates of benign new tasks (Iteration=200) are below 40\%. Thus, we set 40\% as the threshold for detecting adversarial tasks. If the average degradation rate on historical tasks exceeds this threshold, it indicates that the task is adversarial.

\begin{table*}[!t]
	\centering
	\caption{Experimental results. The bolded accuracies are the optimal results, while the underlined ones are sub-optimal results in the same scenario. Due to space limitations, we only listed the average accuracy of the model on all historical data after the training of all tasks, without listing the results of intermediate tasks.}
    \resizebox{\textwidth}{!}{
	\begin{tabular}{c|cccccccc|cc}
    \toprule
          &       & \multicolumn{7}{c|}{\textbf{Continual-based Methods}} & \multicolumn{2}{c}{\textbf{Traditional Methods}} \\
\cmidrule{2-11}          &       & \textbf{$\rm{SacFL}$} & \textbf{CFeD} & \textbf{LwF-Fed} & \textbf{EWC-Fed} & \textbf{MH-Fed} & \textbf{FedWeIT} & \textbf{FCIL} & \textbf{FedAvg} & \textbf{FedProx} \\
    \midrule
    \multirow{9}[1]{*}{\textbf{Class}} & \textbf{FM-3} & \uline{0.64}\tiny{$\pm$2.91e-02}  & 0.1\tiny{$\pm$2.88e-04}   & 0.32\tiny{$\pm$1.81e-02}  & 0.37\tiny{$\pm$2.27e-03}  & 0.48\tiny{$\pm$1.07e-02}  &  0.38\tiny{$\pm$2.23e-03} &  \textbf{0.77}\tiny{$\pm$1.25e-02} & 0.36\tiny{$\pm$2.40e-03}  & 0.1\tiny{$\pm$2.16e-04} \\
          & \textbf{FM-5} &  \textbf{0.43}\tiny{$\pm$5.09e-02}  & 0.1\tiny{$\pm$3.53e-04}   & 0.22\tiny{$\pm$1.67e-03}  & 0.15\tiny{$\pm$5.18e-04}  & 0.28\tiny{$\pm$1.78e-03}  &  0.18\tiny{$\pm$1.19e-02 }  & \uline{0.36}\tiny{$\pm$1.91e-03 }  & 0.25\tiny{$\pm$1.77e-03}  & 0.1\tiny{$\pm$6.24e-18} \\
          & \textbf{C10-3} & 0.3\tiny{$\pm$2.45e-02}   & 0.32\tiny{$\pm$3.78e-03}  & \uline{0.33}\tiny{$\pm$2.86e-03}  & 0.27\tiny{$\pm$8.33e-04}  & \uline{0.33}\tiny{$\pm$1.59e-03}  &   0.25\tiny{$\pm$6.73e-03 } &  \textbf{0.38}\tiny{$\pm$3.79e-03 } & \uline{0.33}\tiny{$\pm$1.29e-03}  & 0.31\tiny{$\pm$3.99e-04} \\
          & \textbf{C10-5} & \textbf{0.47}\tiny{$\pm$9.92e-03}  & 0.14\tiny{$\pm$1.78e-03}  & \uline{0.22}\tiny{$\pm$2.93e-03}  & 0.1\tiny{$\pm$2.10e-05}   & 0.19\tiny{$\pm$3.05e-03}  &  0.17\tiny{$\pm$6.28e-03 }  &  0.18\tiny{$\pm$4.45e-03 } & 0.21\tiny{$\pm$1.27e-03}  & 0.18\tiny{$\pm$4.08e-03} \\
          & \textbf{News-3} & \textbf{0.68}\tiny{$\pm$1.63e-02}  & 0.57\tiny{$\pm$1.22e-03}  & \uline{0.67}\tiny{$\pm$3.06e-03}  & 0.61\tiny{$\pm$2.40e-04}  & 0.65\tiny{$\pm$2.41e-03}  &  0.37\tiny{$\pm$2.36e-03 }  & 0.51\tiny{$\pm$1.81e-03 }  & 0.65\tiny{$\pm$8.13e-04}  & 0.65\tiny{$\pm$1.28e-03} \\
          & \textbf{News-5} & \textbf{0.58}\tiny{$\pm$6.73e-02} & \uline{0.57}\tiny{$\pm$1.07e-02}  & 0.32\tiny{$\pm$2.82e-03}  & 0.46\tiny{$\pm$1.27e-03}  & 0.51\tiny{$\pm$3.06e-02}  &  0.19\tiny{$\pm$7.64e-04 }  & 0.52\tiny{$\pm$1.78e-03 }  & 0.53\tiny{$\pm$6.42e-03}  & \uline{0.57}\tiny{$\pm$3.72e-04} \\
          & \textbf{C100-10} & \textbf{0.32}\tiny{$\pm$2.74e-03}  & 0.04\tiny{$\pm$4.06e-04}  & \uline{0.11}\tiny{$\pm$6.57e-04}  & 0.01\tiny{$\pm$2.25e-05}  & 0.06\tiny{$\pm$4.99e-04}  &  0.05\tiny{$\pm$2.52e-03 }  & 0.09\tiny{$\pm$1.64e-03 }  & 0.08\tiny{$\pm$2.83e-04}  & 0.07\tiny{$\pm$2.49e-04} \\
          & \textbf{C100-15} & \textbf{0.38}\tiny{$\pm$2.94e-03}  & 0.05\tiny{$\pm$1.23e-03}  & 0.09\tiny{$\pm$8.91e-04}  & 0.01\tiny{$\pm$7.80e-19}  & 0.08\tiny{$\pm$4.51e-04}  &  0.05\tiny{$\pm$3.46e-03 }  & \uline{0.10}\tiny{$\pm$7.16e-04 }  & 0.08\tiny{$\pm$1.57e-03}  & 0.08\tiny{$\pm$3.55e-04} \\
          & \textbf{C100-20} & \textbf{0.43}\tiny{$\pm$4.11e-03}  & 0.04\tiny{$\pm$3.09e-04}  & 0.05\tiny{$\pm$3.34e-04}  & 0.01\tiny{$\pm$7.80e-19}  & 0.05\tiny{$\pm$5.37e-04}  & 0.03 \tiny{$\pm$2.24e-03 }  &  \uline{0.06}\tiny{$\pm$1.54e-03 } & 0.05\tiny{$\pm$5.38e-04}  & 0.05\tiny{$\pm$3.77e-04} \\
    \midrule
    \textbf{Domain} & \textbf{C10} & \textbf{0.8}\tiny{$\pm$1.22e-03}   & 0.76\tiny{$\pm$1.62e-03}  & 0.77\tiny{$\pm$1.18e-03}  & 0.76\tiny{$\pm$9.07e-04}  & \uline{0.79}\tiny{$\pm$6.20e-04}  & 0.78\tiny{$\pm$3.10e-03}   & --  & \uline{0.79}\tiny{$\pm$5.20e-04}  & 0.66\tiny{$\pm$9.06e-04} \\
    \bottomrule
    \end{tabular}}%
	\label{results}%
\end{table*}%

\subsection{Simple Class Continual Learning}\label{Simple Continual Learning}

\begin{figure*}[!ht]
	\centering
	\subfloat[Fa-MNIST, task0.]{\includegraphics[width=1.35in]{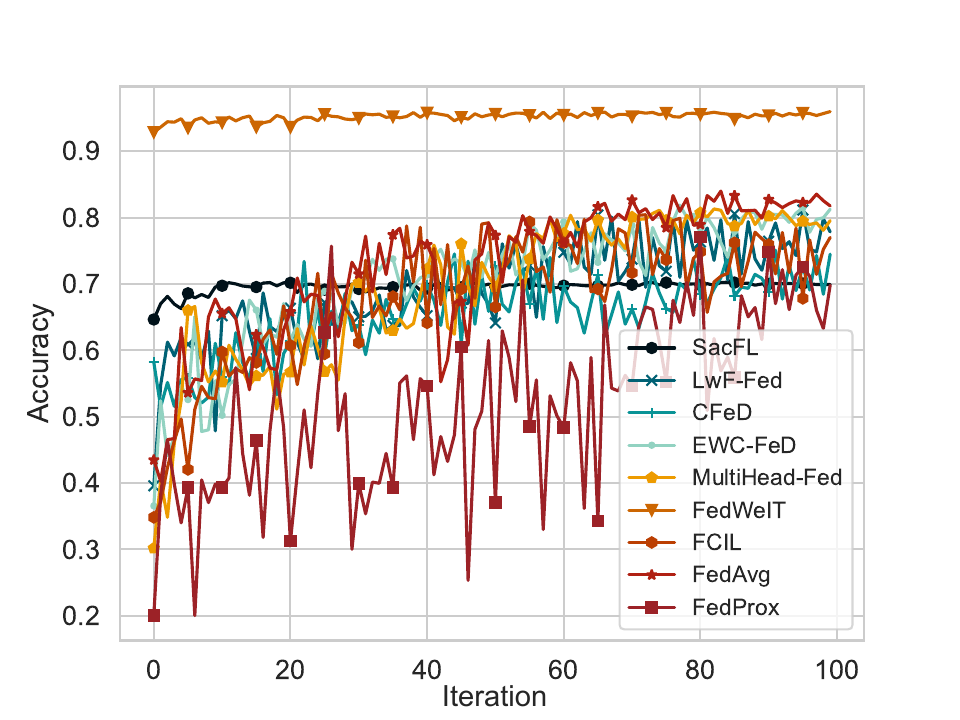}%
		\label{train_acc_model_LeNet_FashionMNIST_tasknum_5_task_0}
	}
	\hfil
	\subfloat[Fa-MNIST, task1.]{\includegraphics[width=1.35in]{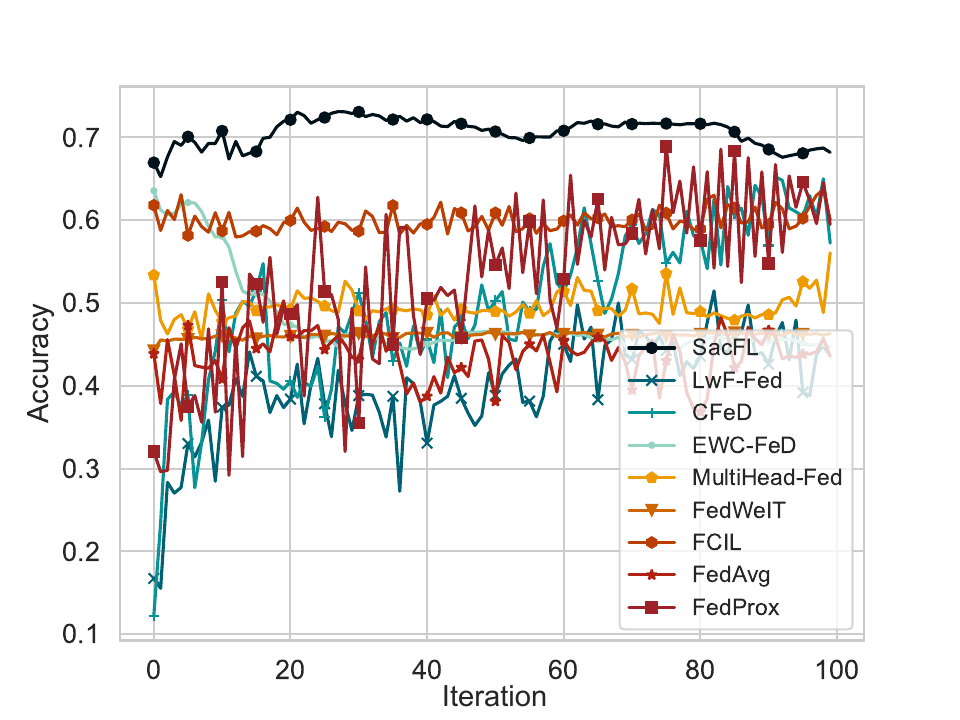}%
		\label{train_acc_model_LeNet_FashionMNIST_tasknum_5_task_1}
	}
	\hfil
	\subfloat[Fa-MNIST, task2.]{\includegraphics[width=1.35in]{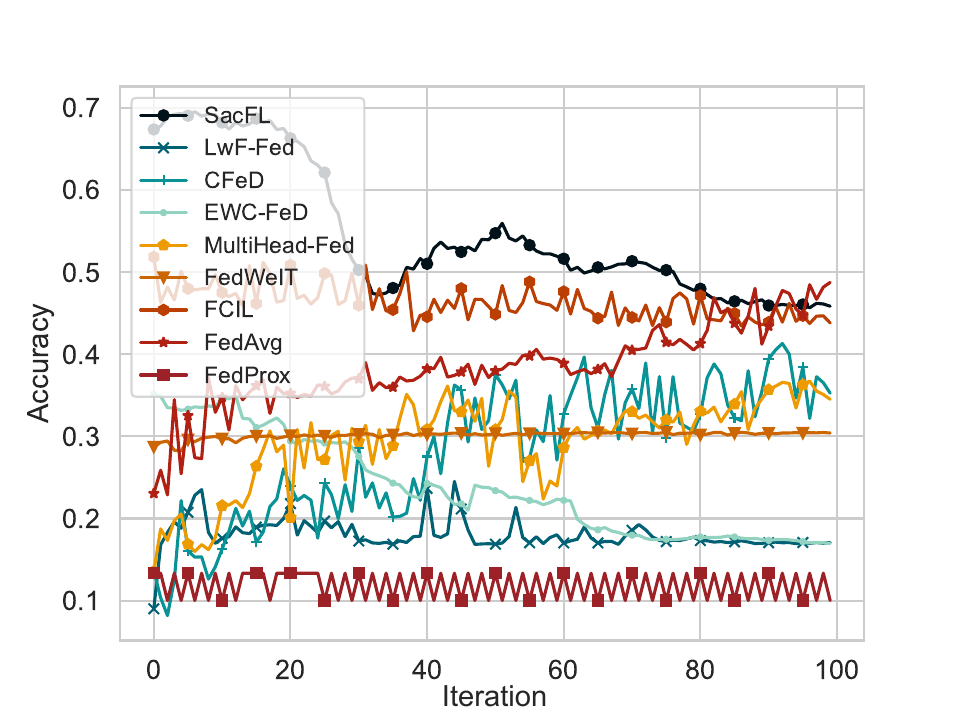}%
		\label{train_acc_model_LeNet_FashionMNIST_tasknum_5_task_2}
	}
	\hfil
	\subfloat[Fa-MNIST, task3.]{\includegraphics[width=1.35in]{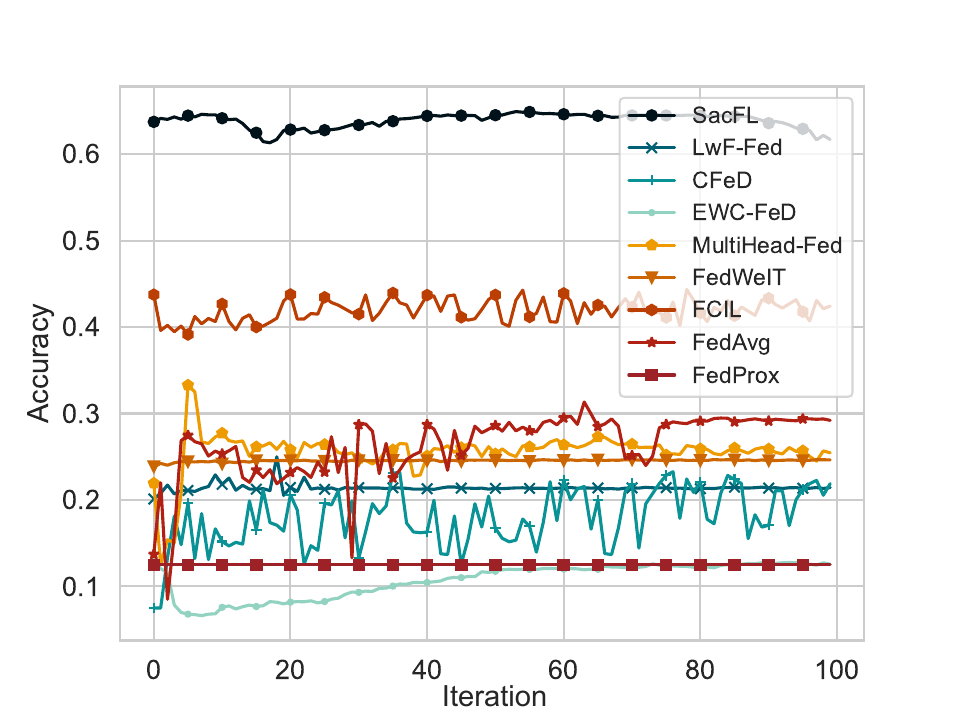}%
		\label{train_acc_model_LeNet_FashionMNIST_tasknum_5_task_3}
	}
	\hfil
	\subfloat[Fa-MNIST, task4.]{\includegraphics[width=1.35in]{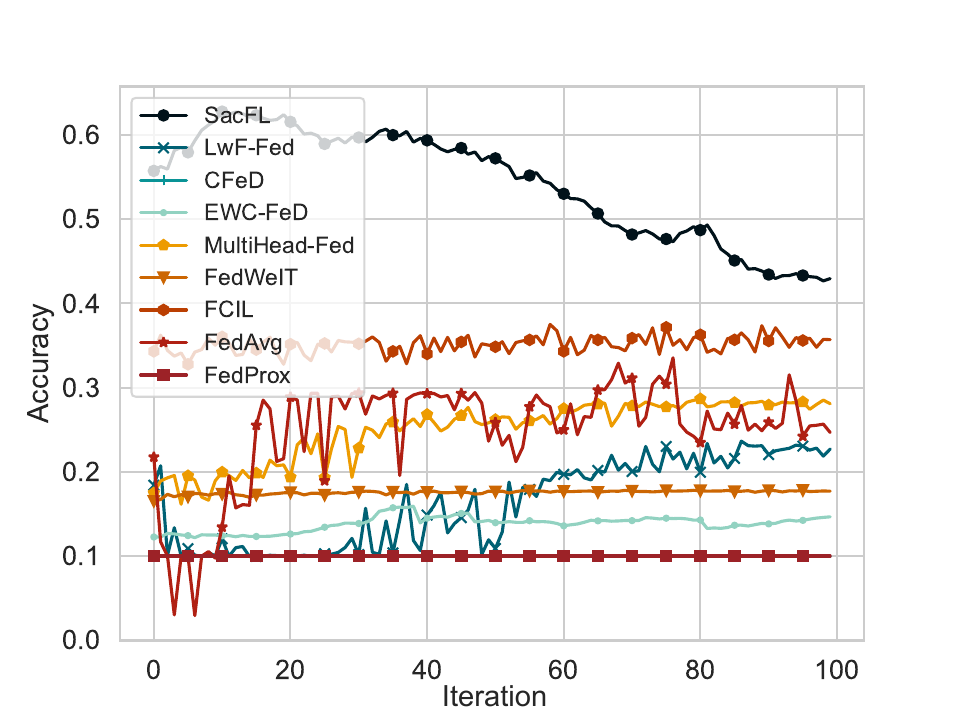}%
		\label{train_acc_model_LeNet_FashionMNIST_tasknum_5_task_4}
	}
	\caption{FashionMNIST, task num=5. }
	\label{fashionmnist_5}
\end{figure*}

\begin{figure*}[!ht]
	\centering
	\subfloat[Cifar10, task0.]{\includegraphics[width=1.4in]{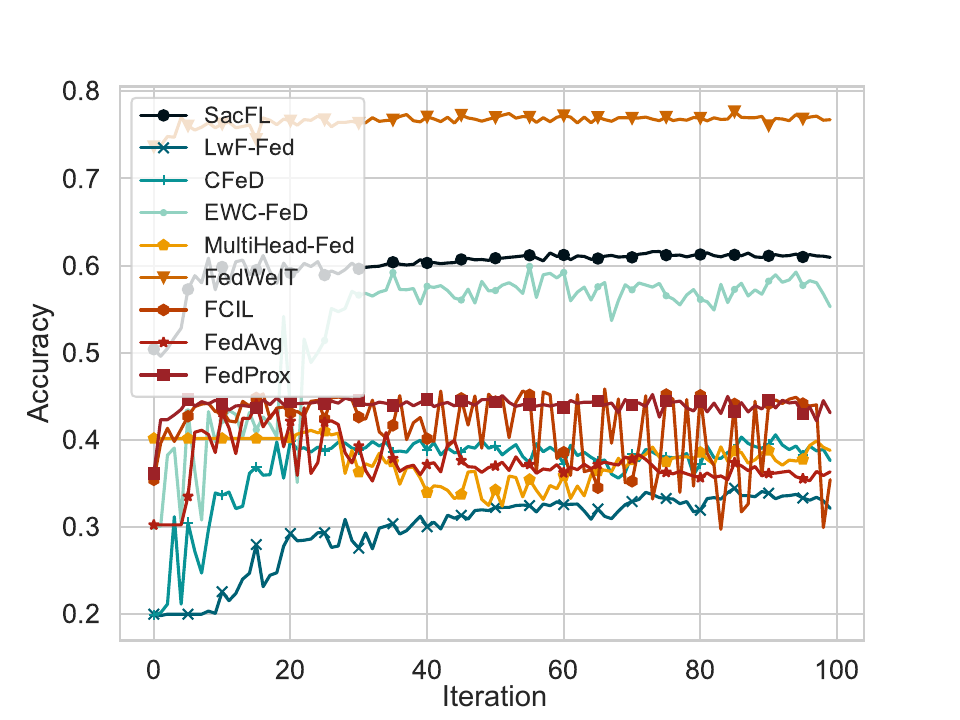}%
		\label{train_acc_model_CNN_Cifar10_tasknum_5_task_0}
	}
	\hfil
	\subfloat[Cifar10, task1.]{\includegraphics[width=1.4in]{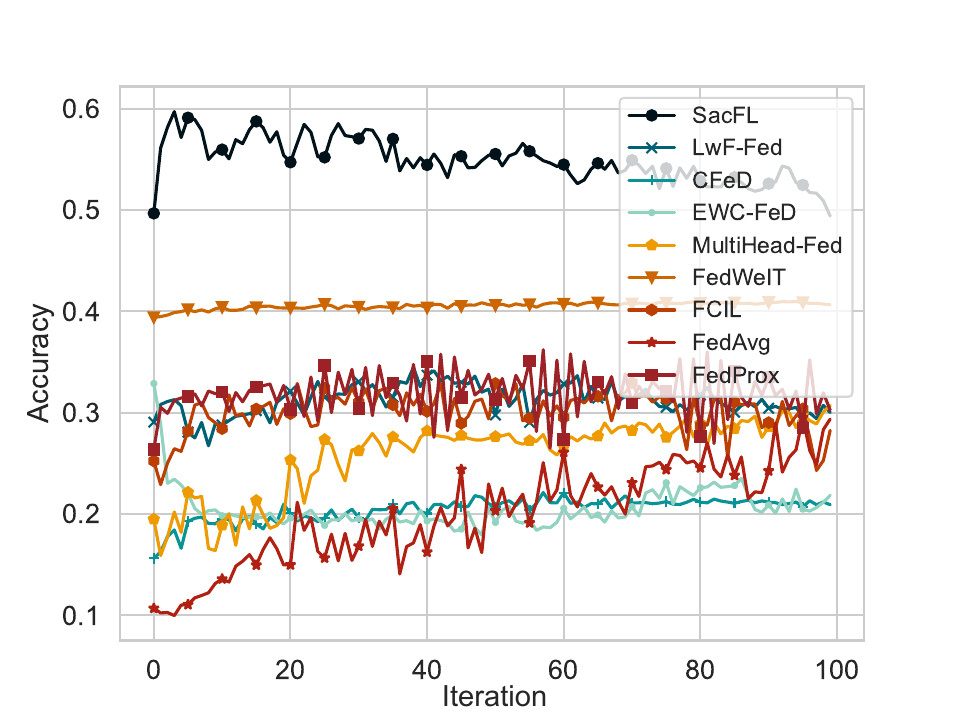}%
		\label{train_acc_model_CNN_Cifar10_tasknum_5_task_1}}
	\hfil
	\subfloat[Cifar10, task2.]{\includegraphics[width=1.4in]{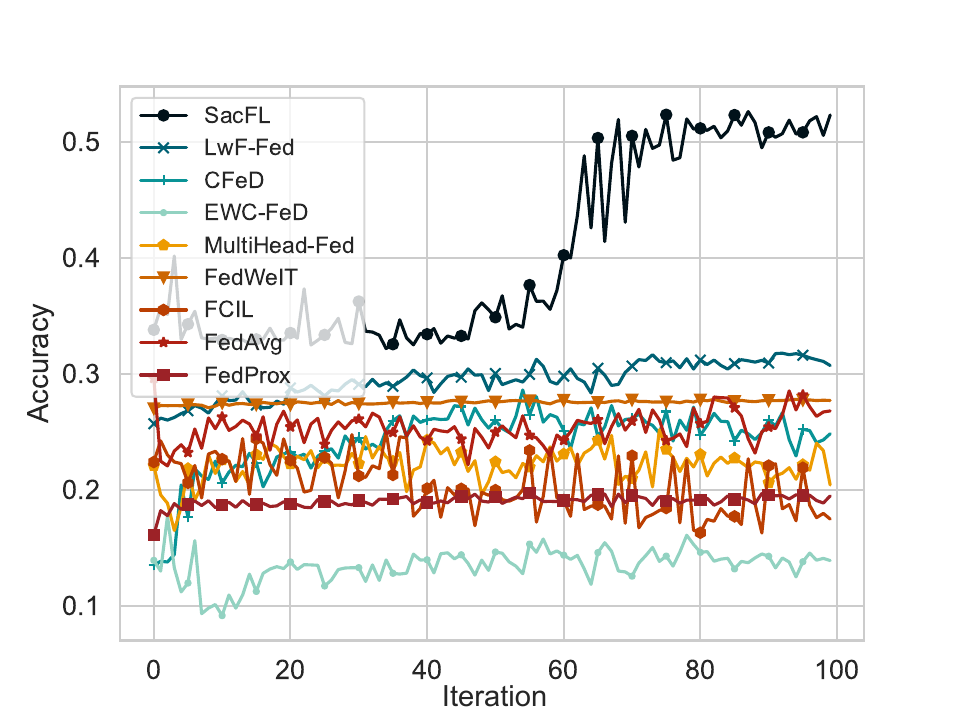}%
		\label{train_acc_model_CNN_Cifar10_tasknum_5_task_2}}
	\hfil
	\subfloat[Cifar10, task3.]{\includegraphics[width=1.4in]{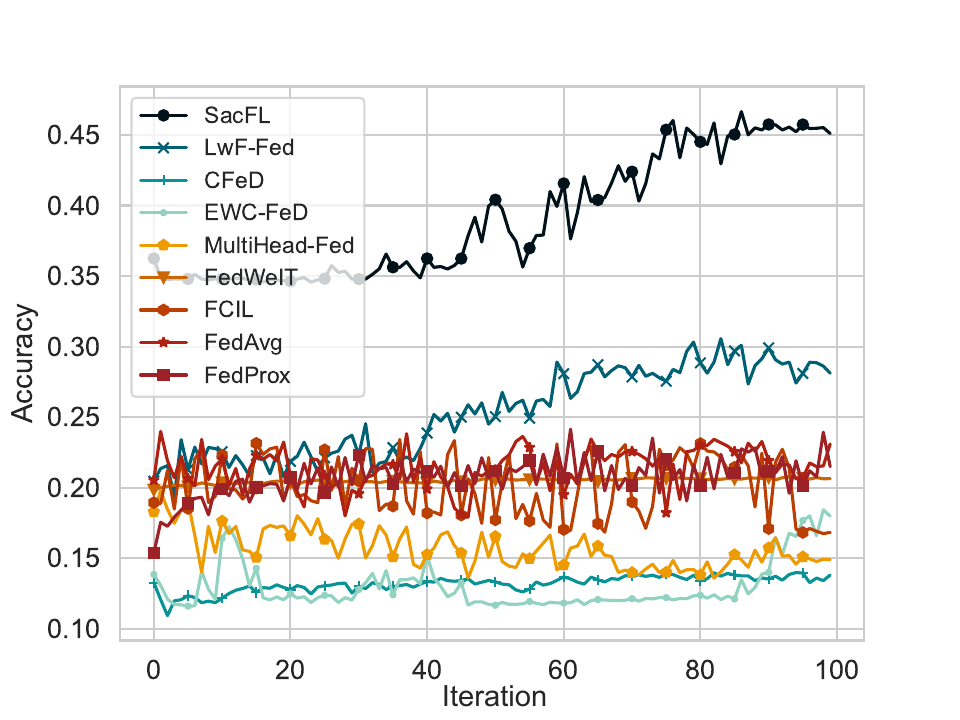}%
		\label{train_acc_model_CNN_Cifar10_tasknum_5_task_3}}
	\hfil
	\subfloat[Cifar10, task4.]{\includegraphics[width=1.4in]{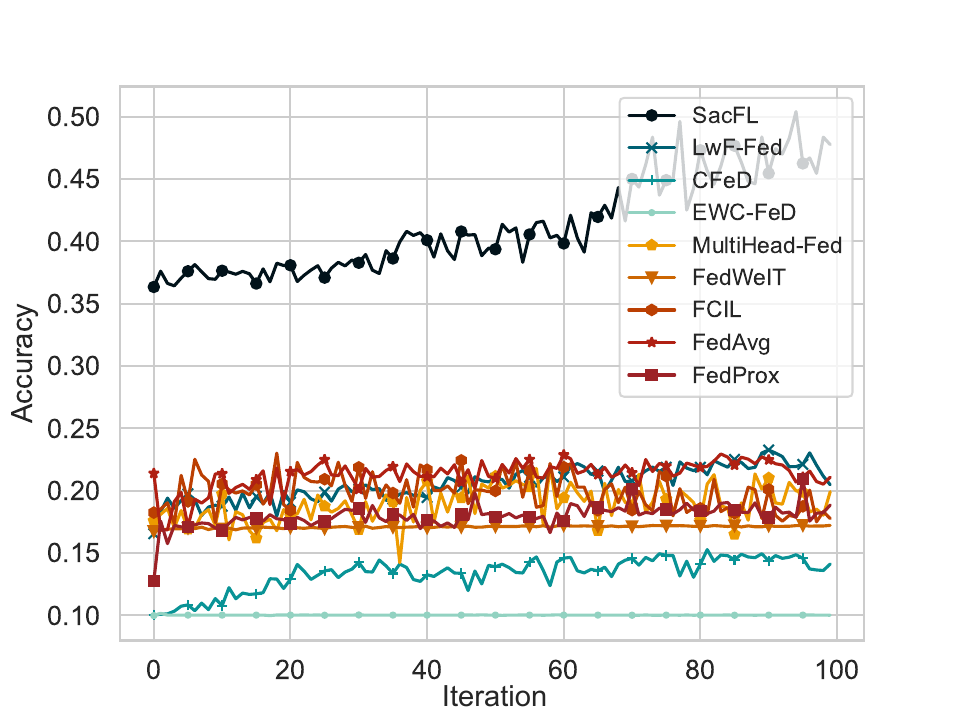}%
		\label{train_acc_model_CNN_Cifar10_tasknum_5_task_4}}	
	\caption{Cifar10, task num=5. }
	\label{cifar10_5}
\end{figure*}

\begin{figure*}[!ht]
	\centering
	\subfloat[THUCNews, task0.]{\includegraphics[width=1.4in]{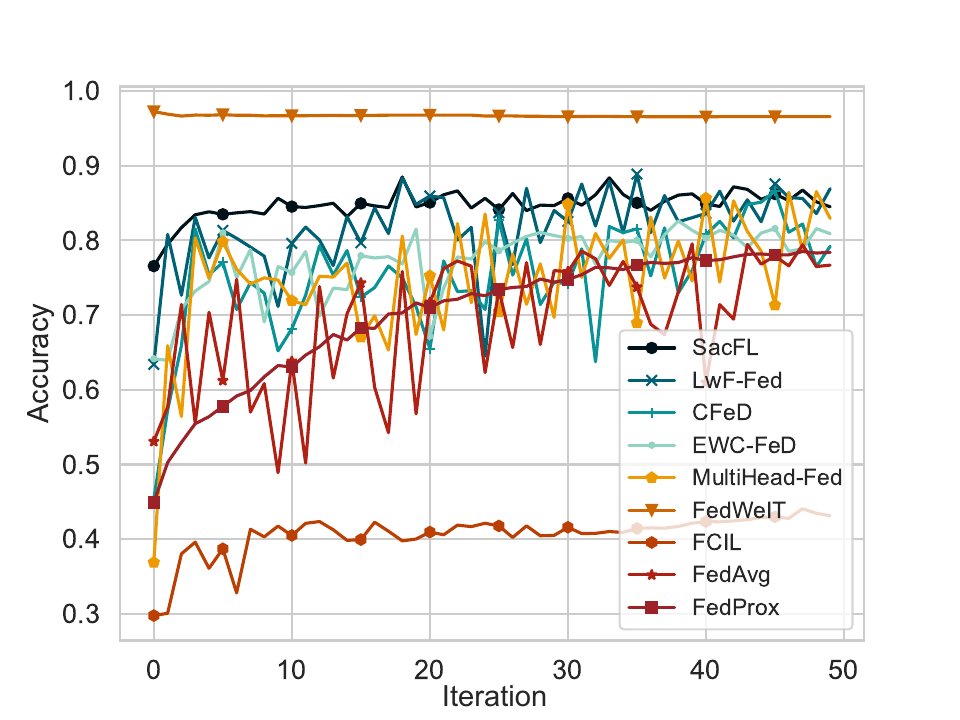}%
		\label{train_acc_model_TextCNN_tasknum_5_task_0}
	}
	\hfil
	\subfloat[THUCNews, task1.]{\includegraphics[width=1.4in]{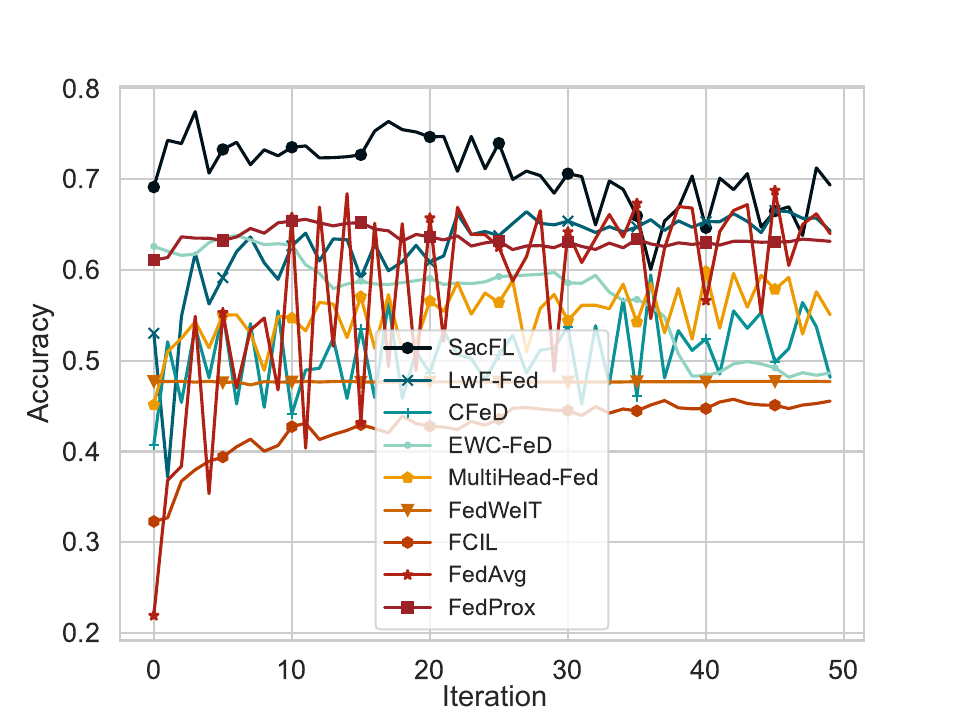}%
		\label{train_acc_model_TextCNN_tasknum_5_task_1}}
	\hfil
	\subfloat[THUCNews, task2.]{\includegraphics[width=1.4in]{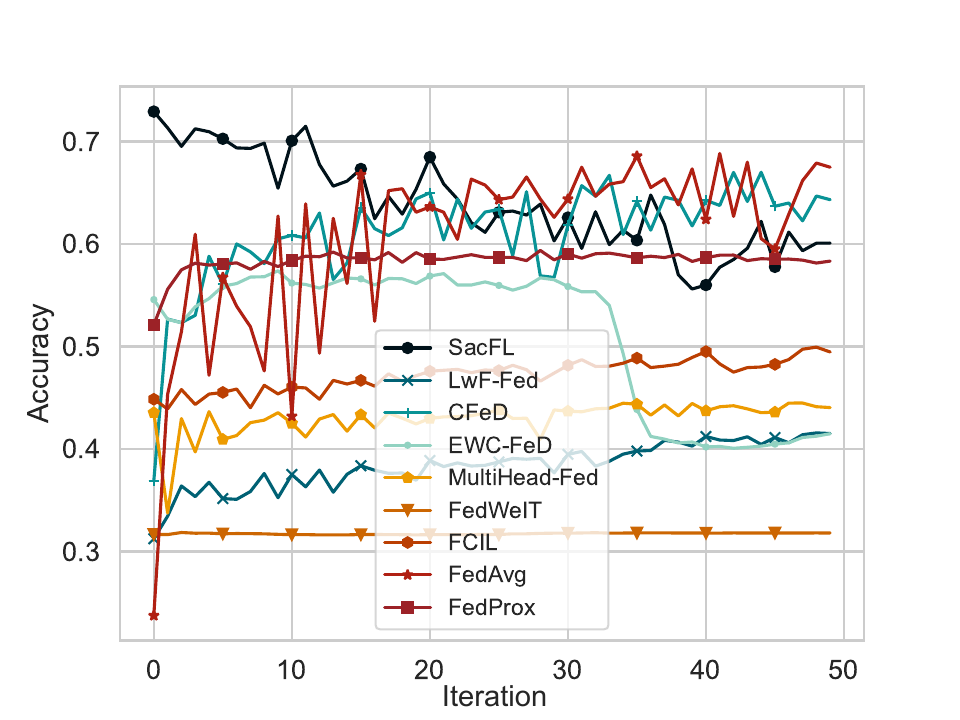}%
		\label{train_acc_model_TextCNN_tasknum_5_task_2}}
	\hfil
	\subfloat[THUCNews, task3.]{\includegraphics[width=1.4in]{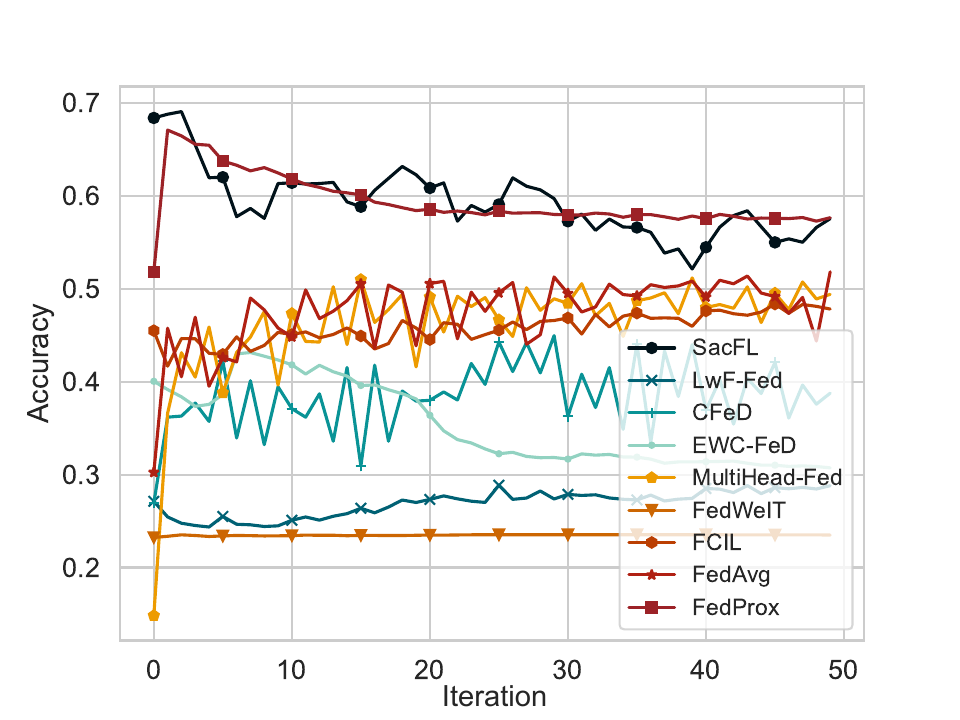}%
		\label{train_acc_model_TextCNN_tasknum_5_task_3}}
	\hfil
	\subfloat[THUCNews, task4.]{\includegraphics[width=1.4in]{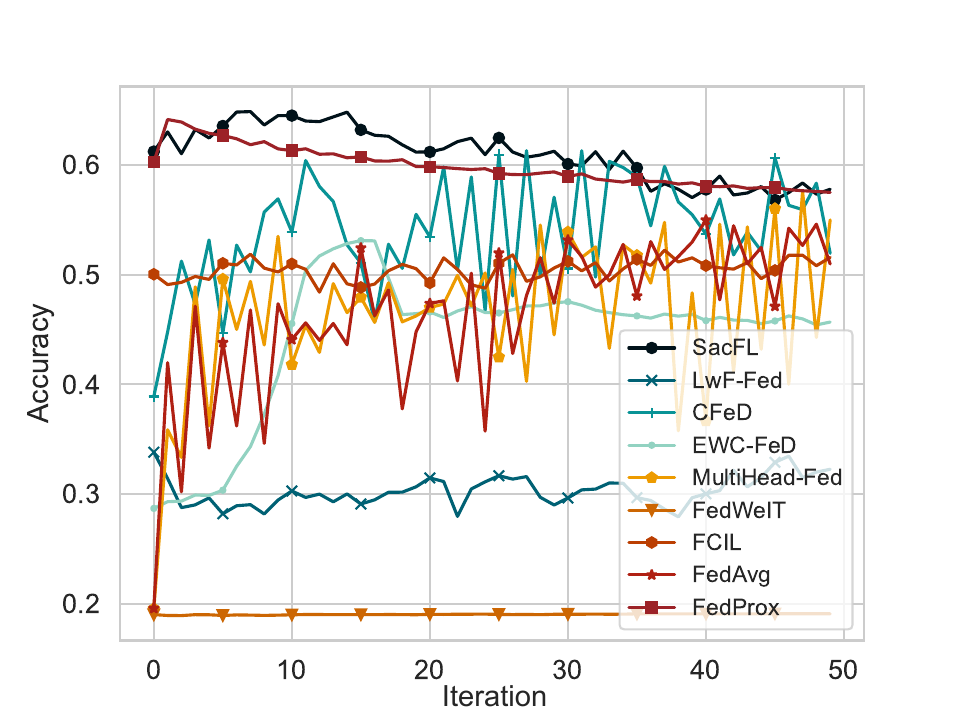}%
		\label{train_acc_model_TextCNN_tasknum_5_task_4}}	
	\caption{THUCNews, task num=5.}
	\label{THUCNews_5}
\end{figure*}

This section focuses on the class-incremental scenario and applies the findings from Section \ref{Data Drift Detection exp} to simple class continual learning using the FashionMNIST, Cifar10, and THUCNews datasets. In this scenario, the data is relatively stable and undergoes shifts only a few times. Therefore, we consider scenarios with 3 and 5 data shifts, corresponding to 3 and 5 tasks, respectively. In the experimental setup, we endeavored to ensure an equal distribution of the class number included in each task. Since each of the aforementioned datasets comprises 10 classes, with 3 tasks, the distribution is as follows: 3, 3, and 4 classes per task, respectively. When there are five tasks, as 10 is divisible by 5, each task contains 2 classes. The experimental results are listed in Tab. \ref{results} and visualized in Fig. \ref{fashionmnist_5}, Fig. \ref{cifar10_5}, and Fig. \ref{THUCNews_5}.

In the results figure, the horizontal axis represents the number of iterations for the current task, while the vertical axis indicates the average accuracy of the model on the testing data from both all historical tasks and the current task. The model for the subsequent task is initialized using the parameters from the previous task. From Tab. \ref{results}, it can be observed that when the total number of tasks is 3, $\rm{SacFL}$ holds an optimal or near-optimal position across various datasets, being on par with most methods, yet it does not demonstrate a distinct advantage. However, when the total number of tasks increases to 5, the accuracy of $\rm{SacFL}$ in tasks 1 to 4 is significantly higher than that of other methods (Fig. \ref{fashionmnist_5}-Fig. \ref{THUCNews_5}), both on the FashionMNIST and Cifar10 datasets. While most methods achieve only 20\%-30\% accuracy in Cifar10, $\rm{SacFL}$ attains 50\%-60\% accuracy. Notably, compared to the scenario with 3 tasks, the advantages of $\rm{SacFL}$ become more pronounced as the number of tasks increases to 5. We speculate that as the number of tasks increases, the superiority of $\rm{SacFL}$ gradually strengthens (verified in Section \ref{Sequential Continual Learning}). The reason behind this is that through monitoring the model layers' changes with tasks, we identify task-sensitive lightweight Decoders and directly leverage the historical information they encapsulate. This ensures the integrity of historical task-related knowledge. Moreover, these lightweight task-sensitive Decoders notably alleviate storage resource demands compared to storing the entire historical model. However, we also observe that in subsequent tasks, while $\rm{SacFL}$ maintains a significant advantage, there may be a slight decline. This is because, on individual datasets, when the average forgetting rate for historical tasks exceeds the learning rate for new tasks, the overall accuracy shows a decreasing trend. The reason behind this is that training in subsequent tasks can introduce minor alterations to the Encoder, diminishing the coupling between the Encoder and Decoders from previous tasks. It is not guaranteed to occur. For example, there is a slight decrease in the FashionMNIST and THUCNews datasets, but a weak upward trend in the Cifar10 and Cifar100 datasets.

\subsection{Sequential Class Continual Learning}\label{Sequential Continual Learning}
\begin{figure*}[!ht]
	\centering
\includegraphics[width=1.0\linewidth]{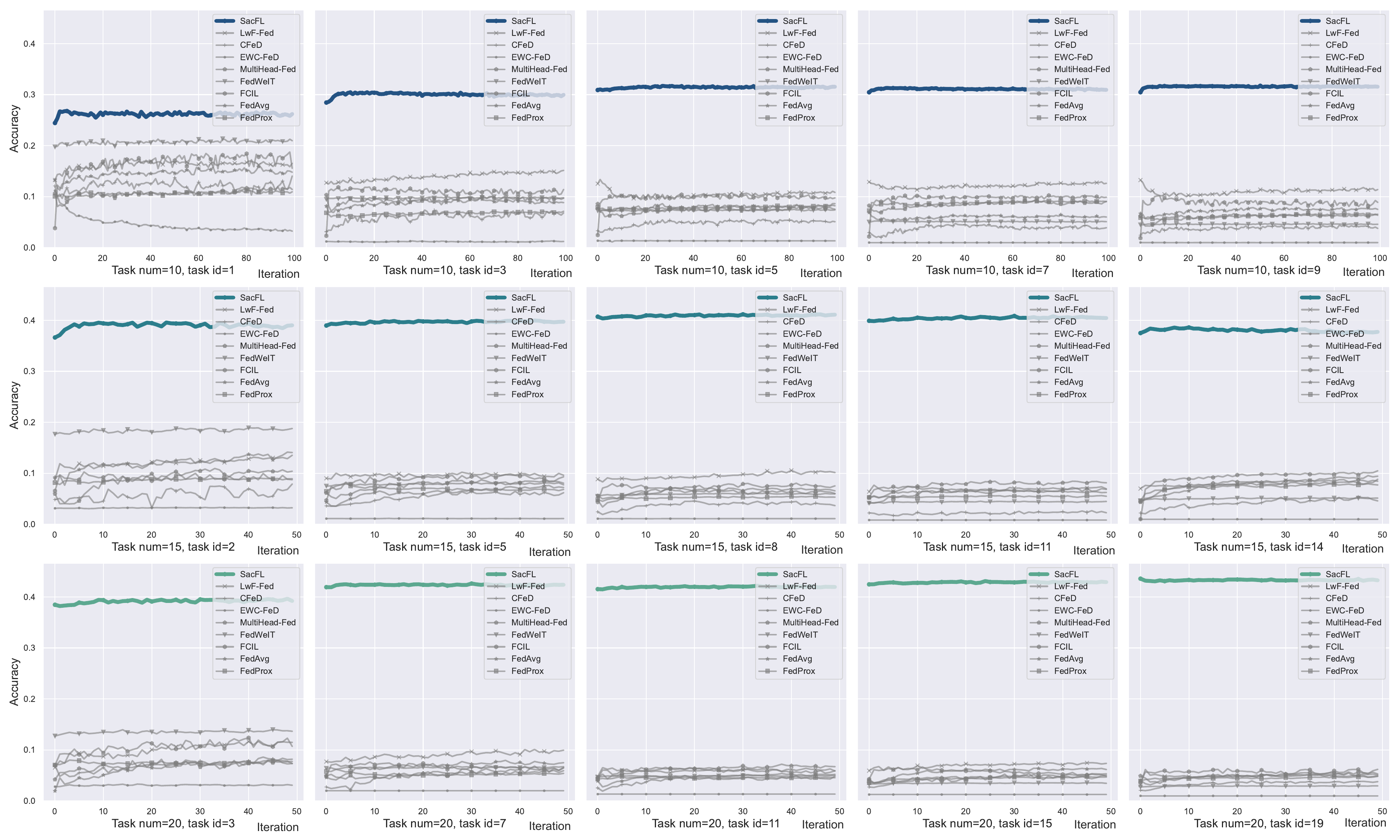}
	\caption{The comparison of performance under Cifar100.}
	\label{Cifar100}
\end{figure*}

In Section \ref{Simple Continual Learning}, we focus on the scenario where the data remains relatively stable, namely simple continual learning. However, in real-world scenarios, continual learning is a long-term endeavor, and the variations across merely 3 or 5 tasks are insufficient. It is necessary to validate $\rm{SacFL}$ in situations with more task variations. Therefore, in this section, we introduce the Cifar100 dataset to construct a larger number of tasks incorporating a wider range of data classes. Our aim is to assess the efficacy of $\rm{SacFL}$ in handling extensive task variations. Specifically, we test the performance under scenarios involving 10, 15, and 20 tasks. Due to space limitations, we only present a subset of the experimental results, as shown in Fig. \ref{Cifar100}.

In Fig. \ref{Cifar100}, the results are depicted for different numbers of tasks: when there are 10 tasks, the outcomes for task 1, task 3, task 5, task 7, and task 9 are displayed; with 15 tasks, the results for task 2, task 5, task 8, task 11, and task 14 are shown; and when there are 20 tasks, the results for task 3, task 7, task 11, task 15, and task 19 are illustrated. It is evident that irrespective of whether the total number of tasks is 10, 15, or 20, traditional methods exhibit minimal effectiveness in sequential tasks, whereas the $\rm{SacFL}$ approach demonstrates a clear advantage and maintains stable convergence. This reaffirms the superior performance of $\rm{SacFL}$ in handling sequential tasks.

\subsection{Domain Continual Learning}\label{Domain Continual Learning}
\begin{figure*}[!ht]
	\centering
	\subfloat[Cifar10, task0.]{\includegraphics[width=0.33\linewidth]{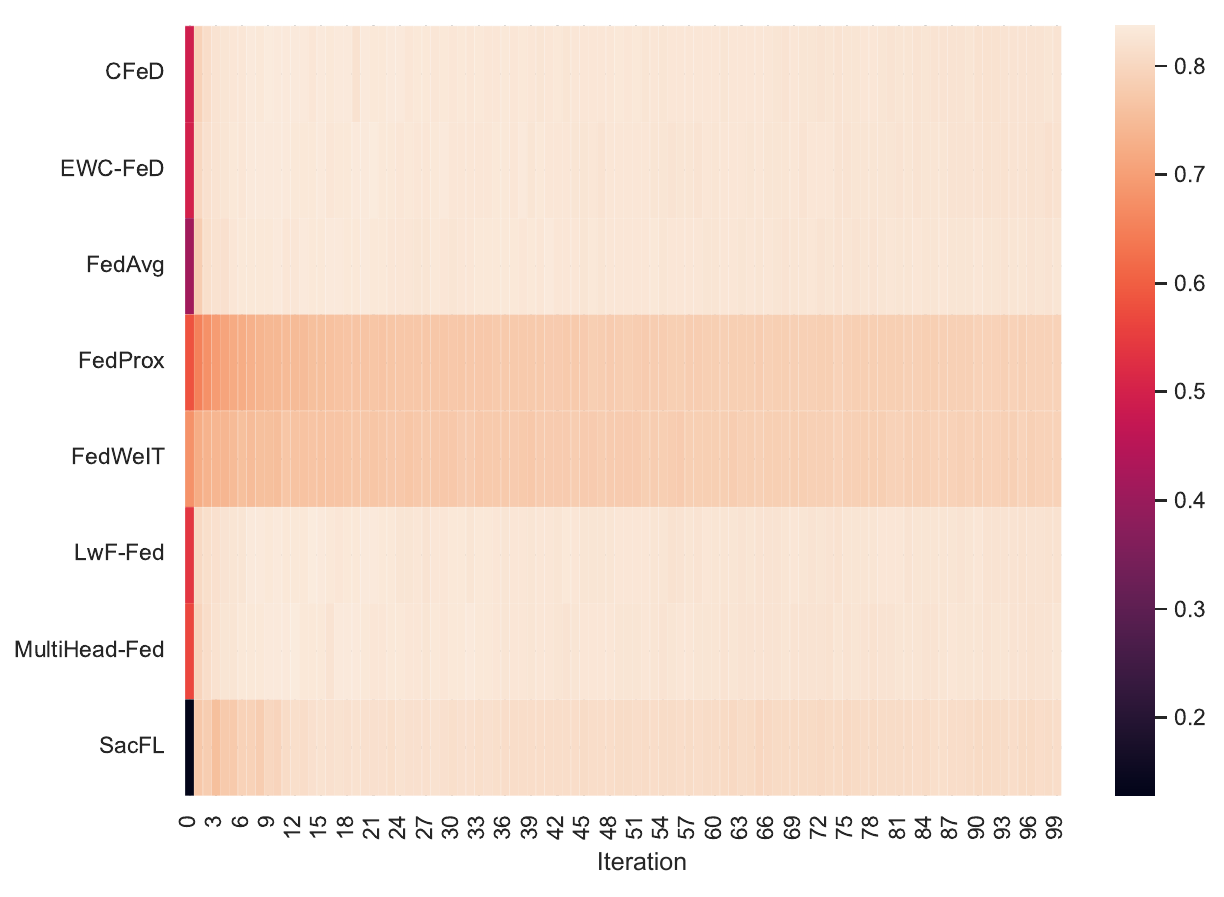}%
		\label{domain_train_acc_model_CNN_Cifar10_tasknum_3_task_0}
	}
	\hfil
	\subfloat[Cifar10, task1.]{\includegraphics[width=0.33\linewidth]{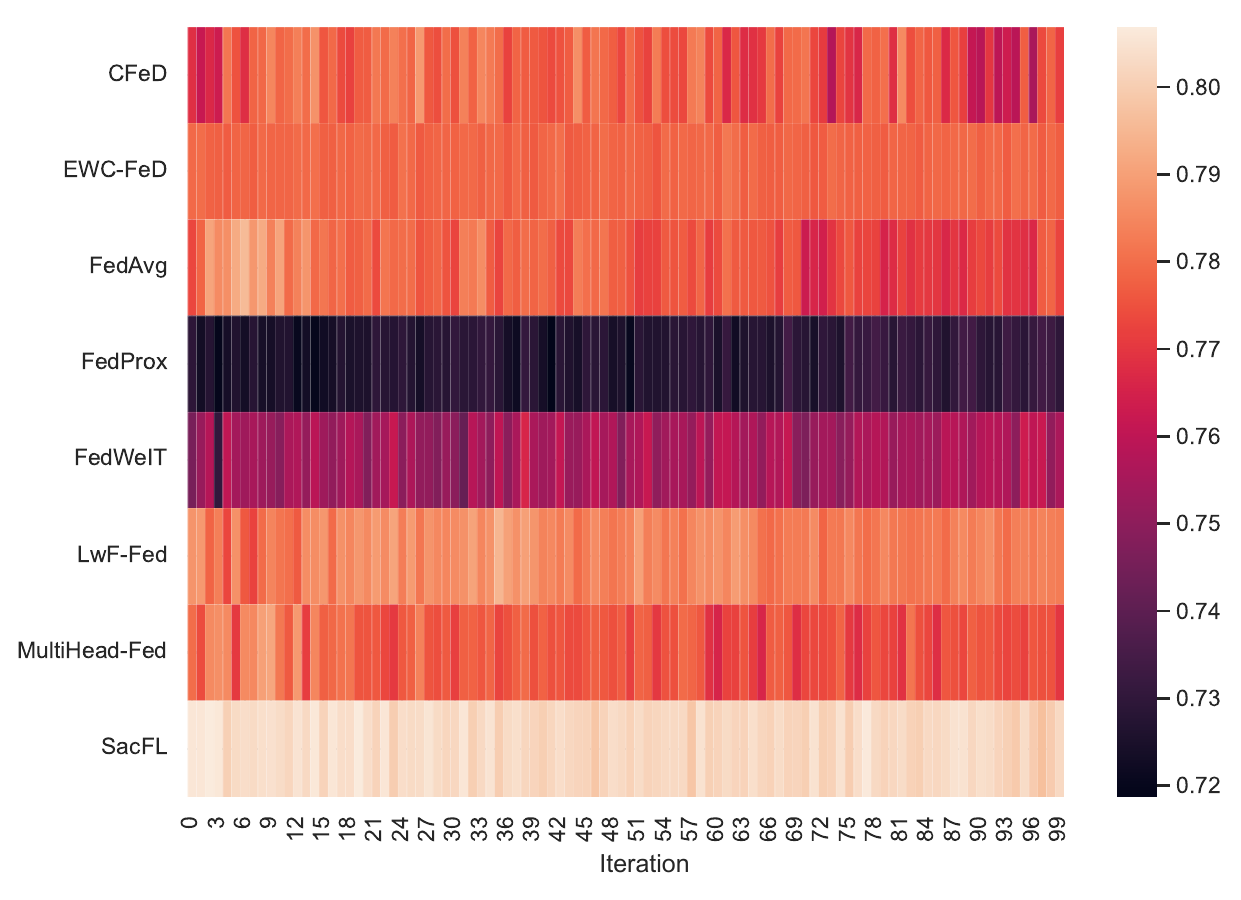}%
		\label{domain_train_acc_model_CNN_Cifar10_tasknum_3_task_1}}
	\hfil
	\subfloat[Cifar10, task2.]{\includegraphics[width=0.33\linewidth]{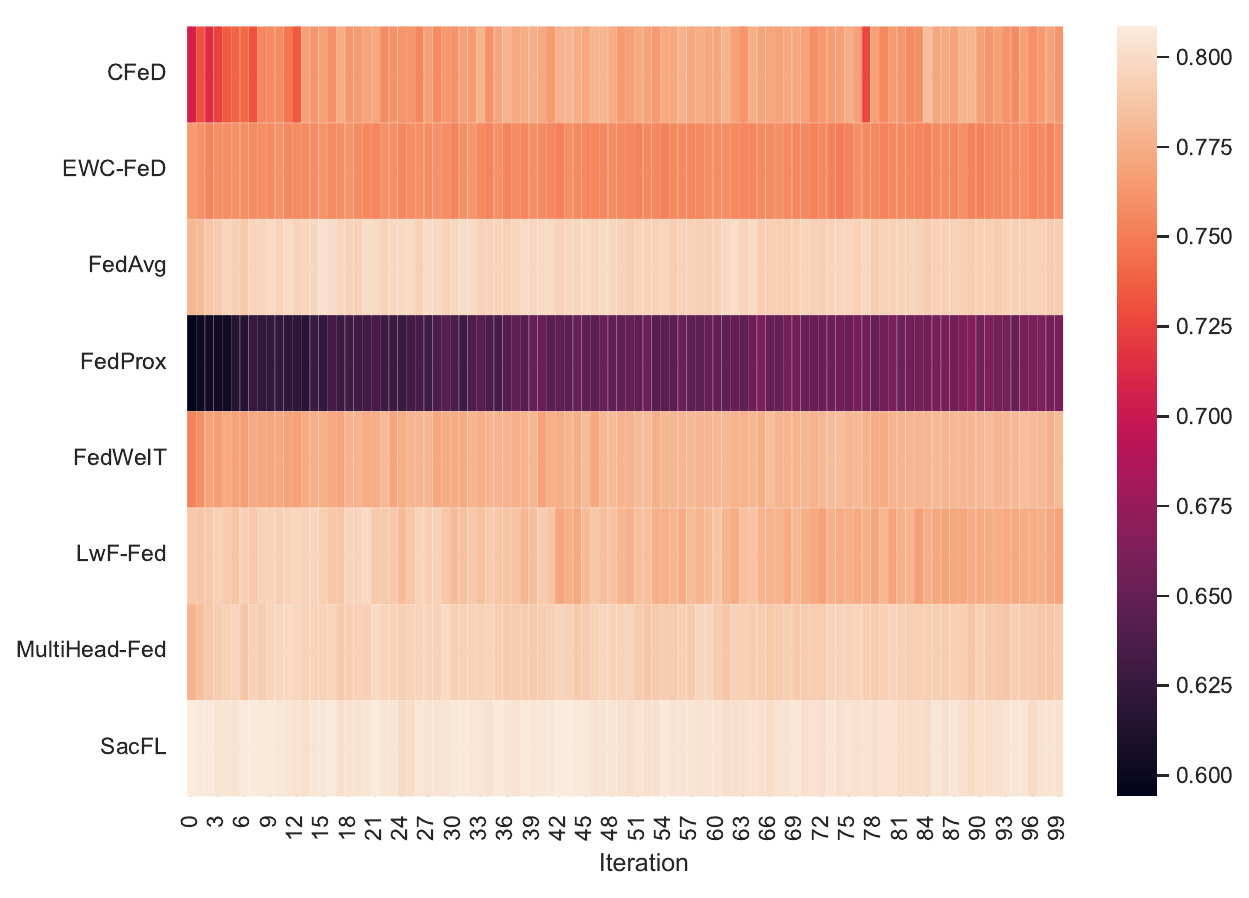}%
		\label{domain_train_acc_model_CNN_Cifar10_tasknum_3_task_2}}
	
	\caption{Cifar10, task num=3, domain-incremental scenario. Each line represents the change in accuracy of a specific algorithm as the number of iterations increases. The lighter the color, the higher the accuracy.}
	\label{domain_cifar10_5}
\end{figure*}

In the experiments in Section \ref{Simple Continual Learning} and Section \ref{Sequential Continual Learning}, validations are carried out under the class-incremental scenario. In addition to class increment, domain increment is also an important setting in continual learning. In the domain-incremental scenario, the labels of the data remain unchanged, but the data itself undergoes shifts. To simulate this scenario, we introduce Gaussian noise and multiplicative noise to the Cifar10 dataset, thus constructing domain-incremental datasets. Consequently, we obtain three tasks: task 0 for the original dataset, task 1 for Gaussian noise, and task 2 for multiplicative noise. The experimental results are illustrated in Fig. \ref{domain_cifar10_5}.

\begin{figure*}[!ht]
	\centering
	\subfloat[Cifar100, task num=5.]{\includegraphics[width=0.33\linewidth]{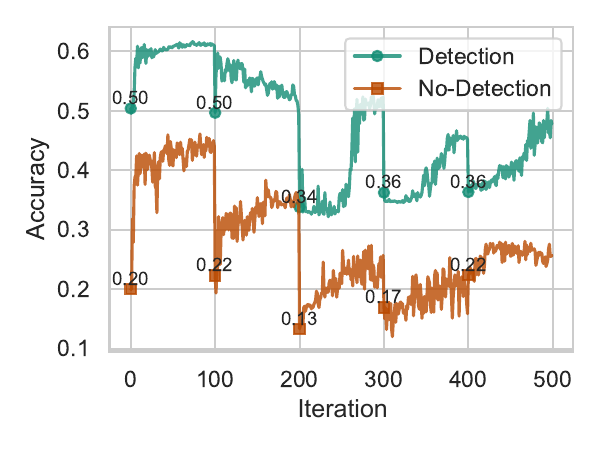}%
		\label{Ablation_fig_CNN_Cifar10_tasknum_5}
	}
	\hfil
	\subfloat[Cifar100, task num=10.]{\includegraphics[width=0.33\linewidth]{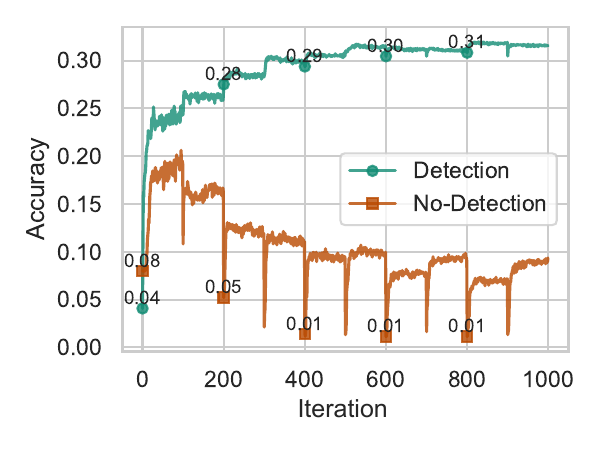}%
		\label{Ablation_fig_CNN_Cifar100_tasknum_10}}
	\hfil
	\subfloat[Cifar100, task num=15.]{\includegraphics[width=0.33\linewidth]{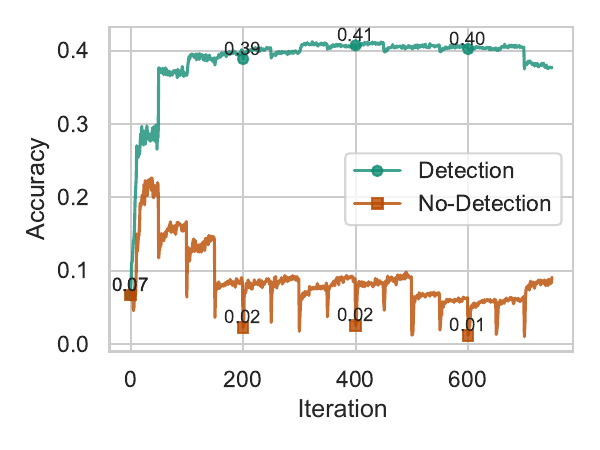}%
		\label{Ablation_fig_CNN_Cifar100_tasknum_15}}
	
	\caption{Ablation study of data shift detection component.}
	\label{Ablation result}
\end{figure*}

In Fig. \ref{domain_cifar10_5}, under the original data (task0), the convergence results and speeds of $\rm{SacFL}$ are consistent with other methods, achieving an accuracy of 80\%. However, upon introducing Gaussian noise to the Cifar10 dataset, all methods exhibit noticeable fluctuations. Except for $\rm{SacFL}$, the performance of other methods significantly decreases. Notably, FedProx, which does not employ continual learning mechanisms, experiences the most significant decline. Furthermore, when the model is further exposed to the multiplicative noise dataset, $\rm{SacFL}$'s accuracy remains high. Therefore, based on the experimental results in Section \ref{Simple Continual Learning}, \ref{Sequential Continual Learning}, and \ref{Domain Continual Learning}, we conclude that $\rm{SacFL}$ performs well in both class-incremental and domain-incremental scenarios in continual learning.

\subsection{Continual Learning under Adversarial Attack}\label{adversarial CL}
All the experiments above are under the assumption that new tasks are benign. However, it is inevitable that some clients are maliciously attacked in the real world. Based on the threshold obtained in Section \ref{Data Drift Detection exp}, we can accurately detect adversarial tasks and trigger the adversarial continual learning mechanism. In this section, we validate our approach using the FashionMNIST and Cifar10 datasets in the contexts of untargeted attacks (label flipping) and targeted attacks (backdoor attacks). The experimental setup assumes a class-incremental learning scenario with 3 tasks, where adversarial data appears in task 1, while Tasks 0 and 3 contain benign samples. The final results are summarized in Tab. \ref{defense performance}. In Tab. \ref{defense performance}, we compare the proposed adversarial continual learning defense method, $\rm{SacFL}$, against commonly used adversarial defense methods in federated learning (Krum \cite{blanchard2017machine}, Median \cite{yin2018byzantine}, and Trimmed\_mean \cite{yin2018byzantine}) under the adversarial task (ID=1) scenario. The reported values represent the test accuracy of the model across all historical tasks. As shown in the table, $\rm{SacFL}$ outperforms other methods overall in terms of defense effectiveness. This demonstrates that $\rm{SacFL}$ is more effective in countering adversarial samples encountered during continual learning.

\begin{table}[htbp]
  \centering
  \caption{Performance of different strategies against attacks.}
  \resizebox{\columnwidth}{!}{
    \begin{tabular}{c|cc|cc}
    \toprule
          & \multicolumn{2}{c|}{\textbf{Label Flipping}} & \multicolumn{2}{c}{\textbf{Backdoor Attack}} \\
\cmidrule{2-5}          & \textbf{Cifar10} & \textbf{F-MNIST} & \textbf{Cifar10} & \textbf{F-MNIST} \\
    \midrule
    \textbf{$\rm{SacFL}$} & \textbf{0.48} \tiny{$\pm$3.35e-02} & \textbf{0.15} \tiny{$\pm$4.71e-02} & \textbf{0.41} \tiny{$\pm$3.32e-02} & \textbf{0.38} \tiny{$\pm$3.81e-02} \\
    \textbf{Krum} & 0.42 \tiny{$\pm$1.86e-02} & 0.11 \tiny{$\pm$3.43e-02} & 0.33 \tiny{$\pm$3.96e-02} & 0.30 \tiny{$\pm$5.88e-02} \\
    \textbf{Median} & 0.43 \tiny{$\pm$2.74e-02} & 0.02 \tiny{$\pm$9.73e-03} & 0.39 \tiny{$\pm$4.02e-02} & 0.14 \tiny{$\pm$3.52e-02} \\
    \textbf{Trim\_m} & 0.47 \tiny{$\pm$1.91e-02} & 0.08 \tiny{$\pm$4.58e-02} & 0.37 \tiny{$\pm$3.21e-02} & 0.15 \tiny{$\pm$3.33e-02} \\
    \bottomrule
    \end{tabular}}%
  \label{defense performance}%
\end{table}%

\subsection{Resource Analysis}\label{Resource Analysis}
When considering the adaptation to limited resources on end devices, $\rm{SacFL}$ demonstrates significant advantages in both computational and storage efficiency compared to other methods, as illustrated in Tab. \ref{tab3}. Especially in the storage aspect, traditional model-based federated continual learning methods typically necessitate storing the entire model to preserve historical knowledge. In contrast, $\rm{SacFL}$ only maintains a lightweight Decoder, thus reducing storage overhead. Taking the ResNet-18 model for Cifar10 as an example, other methods consume 43.73MB/681KB, while the lightweight decoder only occupies 0.19 MB, reducing by 99.9\%; similarly, reductions of 97.7\% for LeNet and 99.9\% for TextCNN. Regarding computation resources, Tab. \ref{tab3} displays the average time consumed per federated iteration when the total number of tasks is 3. We can conclude that compared to the average of other methods, $\rm{SacFL}$ reduces the computing time by 46.22\%, 29.92\%, and 33.33\% for LeNet, ResNet18, and TextCNN, respectively. Therefore, $\rm{SacFL}$ consumes fewer resources overall and is more suitable for end devices with limited resources. It should be noted that the resource consumption of the Multihead method is not listed in the table since it undergoes significant structural changes in each task, making it incomparable to other methods.

\begin{table}[htbp]
  \centering
  \caption{The computation (denoted by C) and storage (denoted by S)overhead of different continual learning methods.}
  \resizebox{\columnwidth}{!}{
    \begin{tabular}{ccccccrc}
    \toprule
          &       & \textbf{$\rm{SacFL}$} & \textbf{CFeD} & \textbf{LwF-Fed} & \textbf{EWC-Fed} & \multicolumn{1}{c}{\textbf{FCIL}} & \textbf{FedWeIT} \\
    \midrule
    \multirow{2}[2]{*}{LeNet} & S     & \textbf{4K} & 177K & 177K & 177K & \multicolumn{1}{c}{177K} & \uline{171K} \\
          & C     & \uline{5.02}  & 5.22  & \textbf{4.92} & 7.18  & \multicolumn{1}{c}{9.91} & 19.44 \\
\cmidrule{2-8}    \multirow{2}[2]{*}{Resnet18} & S     & \textbf{21K} & 43.73M & 43.73M & 43.73M & \multicolumn{1}{c}{43.73M} & \uline{681K} \\
          & C     & \uline{21.79} & 23.08 & \textbf{18.75} & 27.7  &   23.46    & 62.5 \\
\cmidrule{2-8}    \multirow{2}[2]{*}{TextCNN} & S     & \textbf{5K} & 10.51M & 10.51M & 10.51M & \multicolumn{1}{c}{10.51M} & \uline{301K} \\
          & C     & \textbf{3.42} & \uline{4.08}  & 4.34  & 4.43  &    6.57   & 6.23 \\
    \bottomrule
    \end{tabular}}%
  \label{tab3}%
\end{table}%

\subsection{Ablation Studies}\label{Ablation Studies}

In this section, we perform ablation validation on the data drift detection component. Due to space constraints, we specifically focus on validation within the class-incremental scenario involving a large number of classes, yielding results as depicted in Fig. \ref{Ablation result}. It can be observed that in the absence of data drift detection, the model's performance deteriorates with task transitions. However, upon integrating the data drift detection component, the model's performance just experiences only a brief decline after task changes, yet it recovers during subsequent training.

\subsection{Demo System}\label{Real Machine Experiments}
\begin{figure}[!ht]
	\centering
	\includegraphics[width=1.0\linewidth]{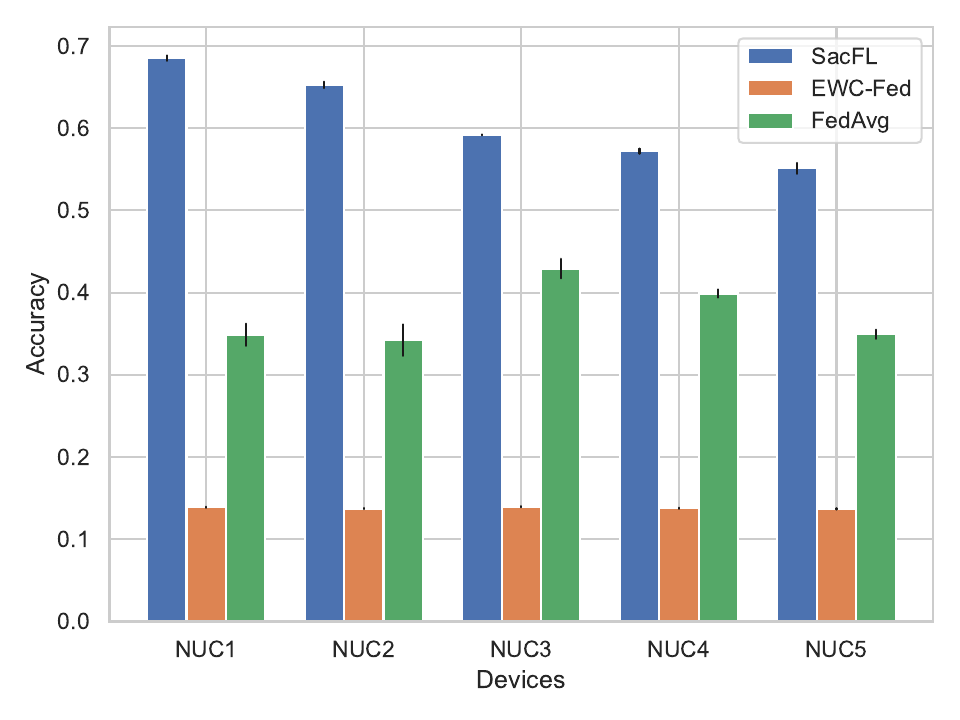}
	\caption{The performance of $\rm{SacFL}$ on the demo system.}
	\label{proto performance}
\end{figure}

In addition to validating $\rm{SacFL}$ in a simulation system, we also develop a distributed demo system, consisting of 5 mini computers NUC with CPU and a central server. The NUCs are equipped with Intel(R) Core(TM) i7-10710U processors, 24GB of RAM, and run on Ubuntu 18.04. The central server contains 4 NVIDIA GeForce RTX 3090 GPUs, and 128GB of RAM, and operates on Ubuntu 22.04. All the NUCs are connected through the lEEE 802.11 wireless network. Leveraging the FashionMNIST dataset, we compare the performance of $\rm{SacFL}$ with that of typical continual learning methods such as EWC-Fed and the non-continual learning method FedAvg, as depicted in Fig. \ref{proto performance}. In Fig. \ref{proto performance}, the test results on all historical tasks for the 5 NUCs are presented after training. It can be observed that the $\rm{SacFL}$ model exhibits overwhelming superiority over the other two methods across all clients. Therefore, $\rm{SacFL}$ maintains its advantage in realistic distributed computing scenarios.

\section{Conclusion}
This paper addresses the problem of continual learning for resource-constrained end devices, proposing a federated continual learning method called $\rm{SacFL}$. $\rm{SacFL}$ identifies that the last few layers are highly sensitive to task variations. Based on this observation, the model is divided into a task-robust Encoder and a task-sensitive lightweight Decoder. By only storing the lightweight Decoders instead of the whole model or historical data on end devices, the overhead of storage and computation resources can be effectively reduced. Moreover, a data shift detection mechanism based on contrastive learning is introduced to detect task changes. It can autonomously identify new tasks and determine whether they are adversarial. For benign tasks, it triggers the CL mechanism, while for adversarial tasks, it activates the attack-defense strategy. Experimental validations conducted on both image and text datasets yield five key conclusions: (1) $\rm{SacFL}$ demonstrates advantages over mainstream continual learning and conventional methods, particularly evident when encountering more frequent changes. (2) $\rm{SacFL}$ greatly reduces the storage and computing overhead on end devices, achieving a reduction ratio of up to 99.9\%, especially in terms of storage resources. (3) Beyond class-incremental scenarios, $\rm{SacFL}$ remains effective in domain-incremental scenarios. (4) In scenarios where the new task is malicious, its effectiveness in mitigating attacks exceeds that of common federated robust aggregation methods. (5) Except for the simulation system, $\rm{SacFL}$ is also effective in a real demo system, demonstrating its practicality.

\normalem
\bibliographystyle{ieeetr}
\bibliography{ref}
\vfill

\begin{IEEEbiography}[{\includegraphics[width=1in,height=1.25in,clip,keepaspectratio]{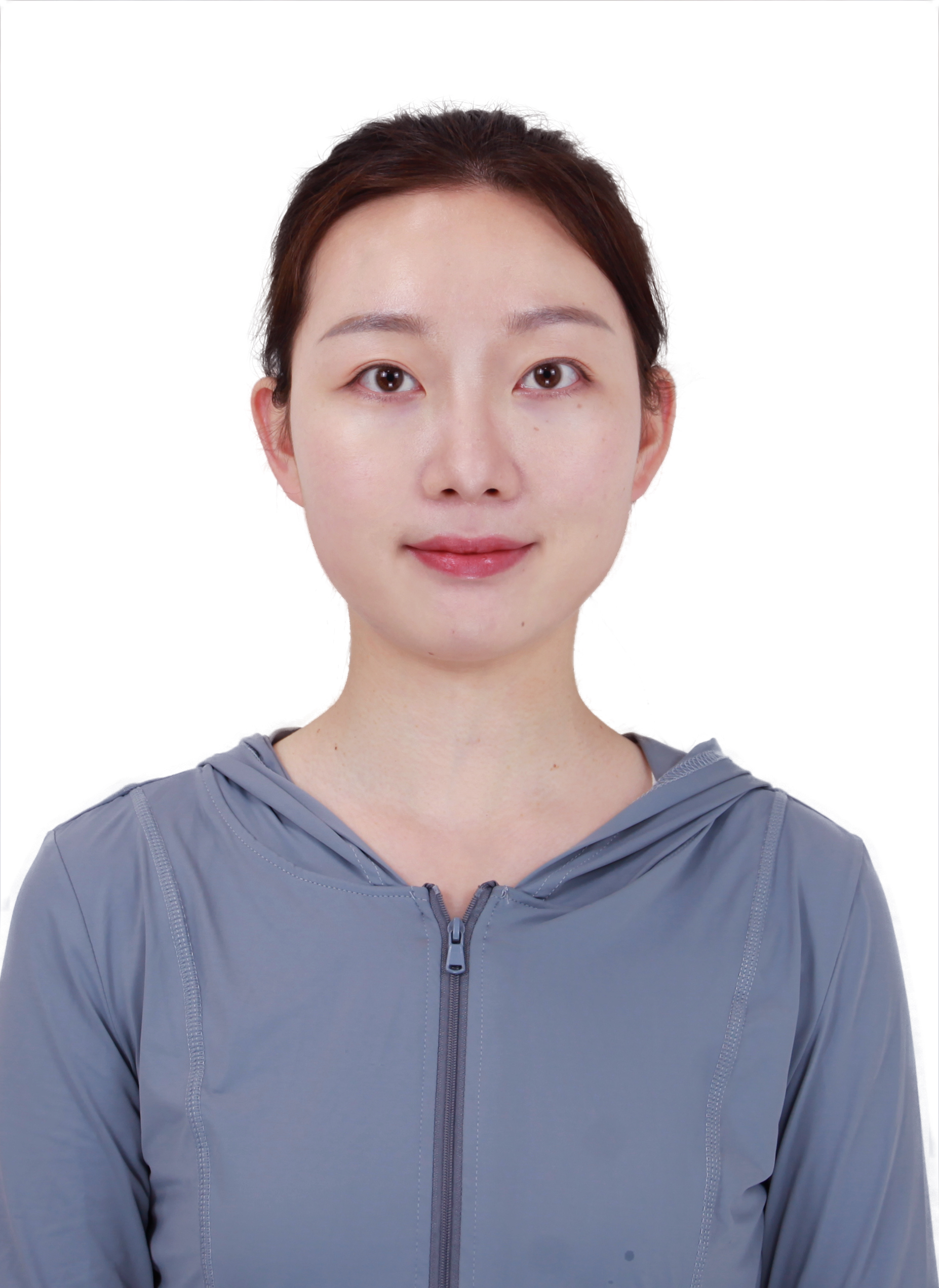}}]{Zhengyi Zhong}
received the B.S. degree from the College of Systems Engineering, National University of Defense Technology, Changsha, China, in 2020, where she is currently pursuing the Ph.D. degree. Her research interests include federated learning, continual learning, machine unlearning, and domain adaptation.
\end{IEEEbiography}

\begin{IEEEbiography}[{\includegraphics[width=1in,height=1.25in,clip,keepaspectratio]{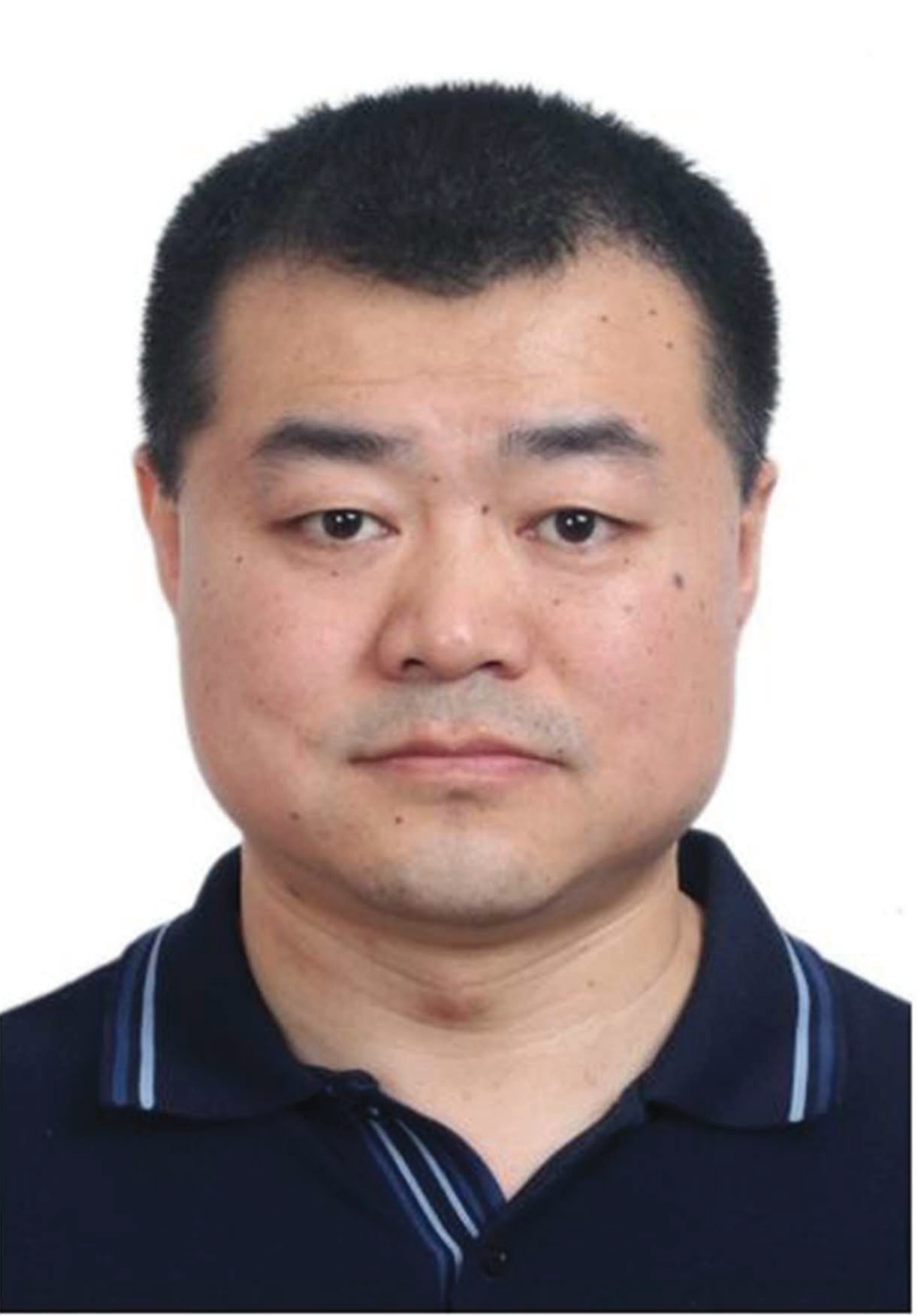}}]{Weidong Bao}
received the Ph.D. degree in information system from the National University of Defense Technology, Changsha, China, in 1999. He is currently a Professor at the College of Systems Engineering, National University of Defense Technology. He has authored more than 100 research articles in refereed journals and conference proceedings, such as IEEE-TC, IEEE-TPDS, IEEE-IoTJ. His recent research interests include cloud computing, information systems, and complex networks.
\end{IEEEbiography}

\begin{IEEEbiography}[{\includegraphics[width=1in,height=1.25in,clip,keepaspectratio]{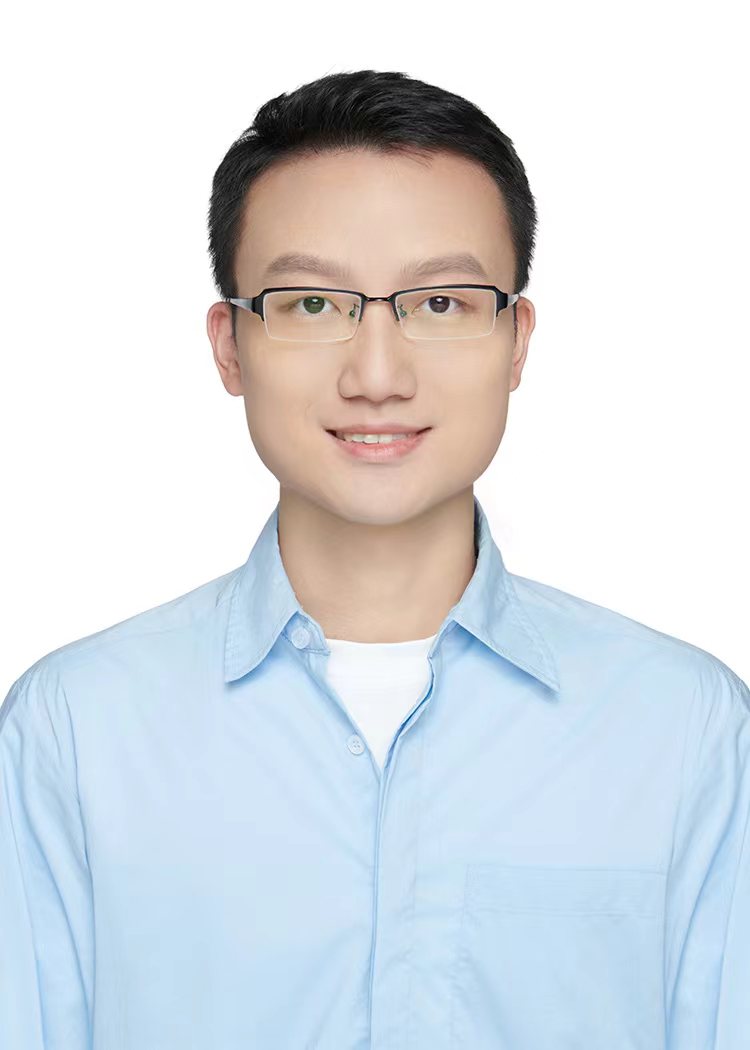}}]{Ji Wang}
received the Ph.D. degree in information system from the National University of Defense Technology, Changsha, China, in 2019. He was a visiting Ph.D. student with the University of Illinois at Chicago, Chicago, IL, USA, from March 2017 to September 2018, under the supervision of Prof. Philip S. Yu. He is currently an Associate Professor with the College of Systems Engineering, National University of Defense Technology. He has authored more than 30 research articles in refereed journals and conference proceedings, such as IEEE-TC, IEEE-TPDS, SIGKDD and AAAI. His research interests include deep learning and edge intelligence.
\end{IEEEbiography}

\begin{IEEEbiography}[{\includegraphics[width=1in,height=1.25in,clip,keepaspectratio]{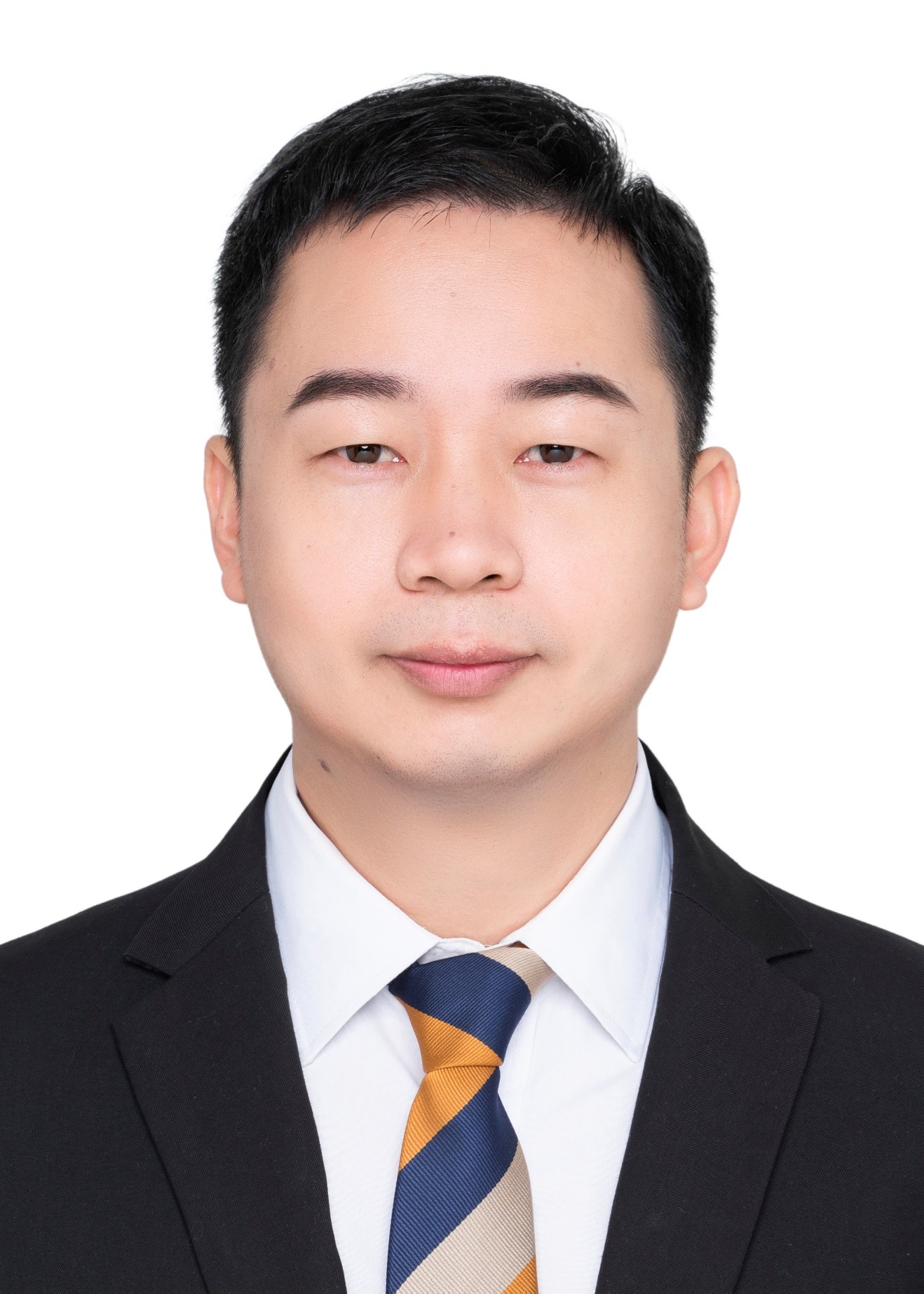}}]{Jianguo Chen}
received his Ph.D. degree in Computer Science and Technology from Hunan University. He is currently an Associate Professor and one of the Hundred Academic Talents in the School of Software Engineering of Sun Yat-sen University (SYSU). He has published more than 70 research papers in international conferences and journals such as IEEE-TII, IEEE-TITS, IEEE-TPDS, IEEEE-TKDE, and IEEE/ACM-TCBB. His major research interests include high-performance artificial intelligence, federated learning, distributed computing, and the application in intelligent transportation and intelligent medicine.
\end{IEEEbiography}

\begin{IEEEbiography}[{\includegraphics[width=1in,height=1.25in,clip,keepaspectratio]{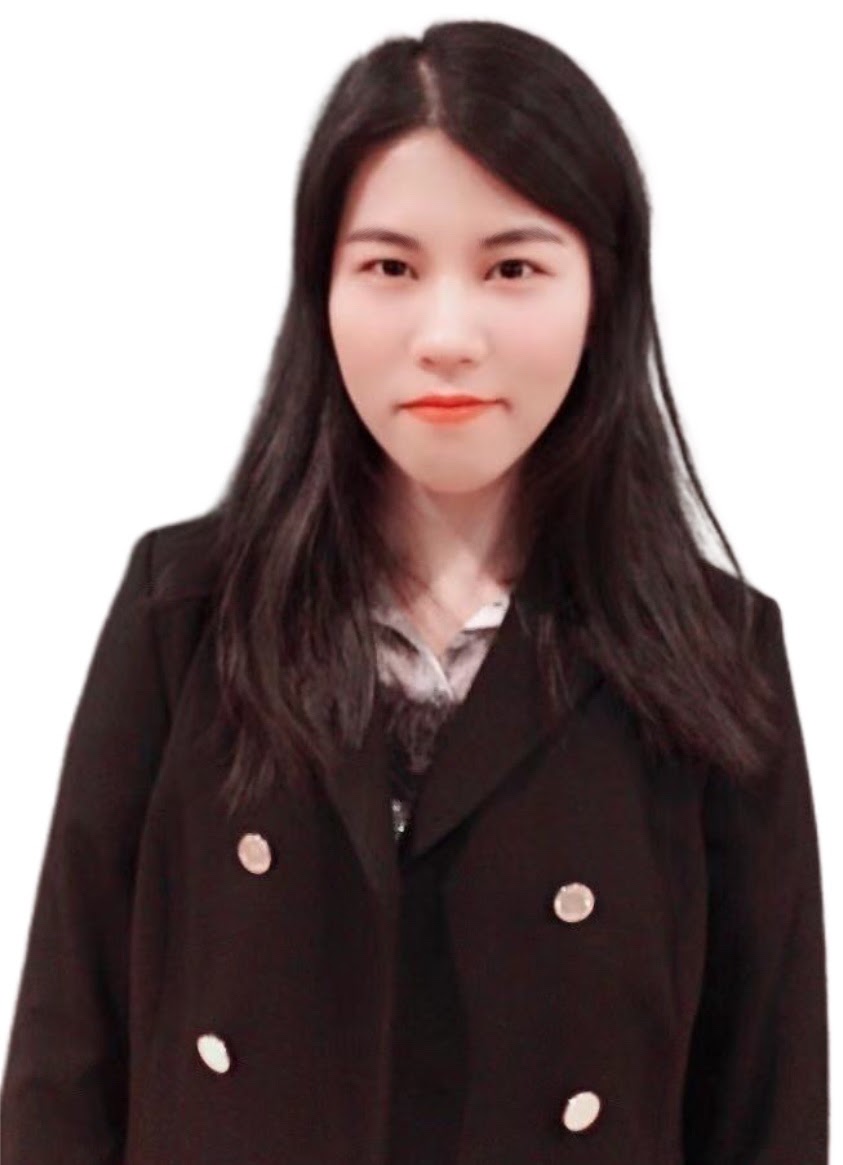}}]{Lingjuan Lyu}
received Ph.D. degree from the University of Melbourne, Melbourne, VIC, Australia, in 2018. She is currently a Research Fellow with National University of Singapore. She was a Research Fellow (Level B3) with Australian National University. Her current research interests include federated learning, trustworthy AI, edge intelligence, and fairness.
\end{IEEEbiography}

\begin{IEEEbiography}[{\includegraphics[width=1in,height=1.25in,clip,keepaspectratio]{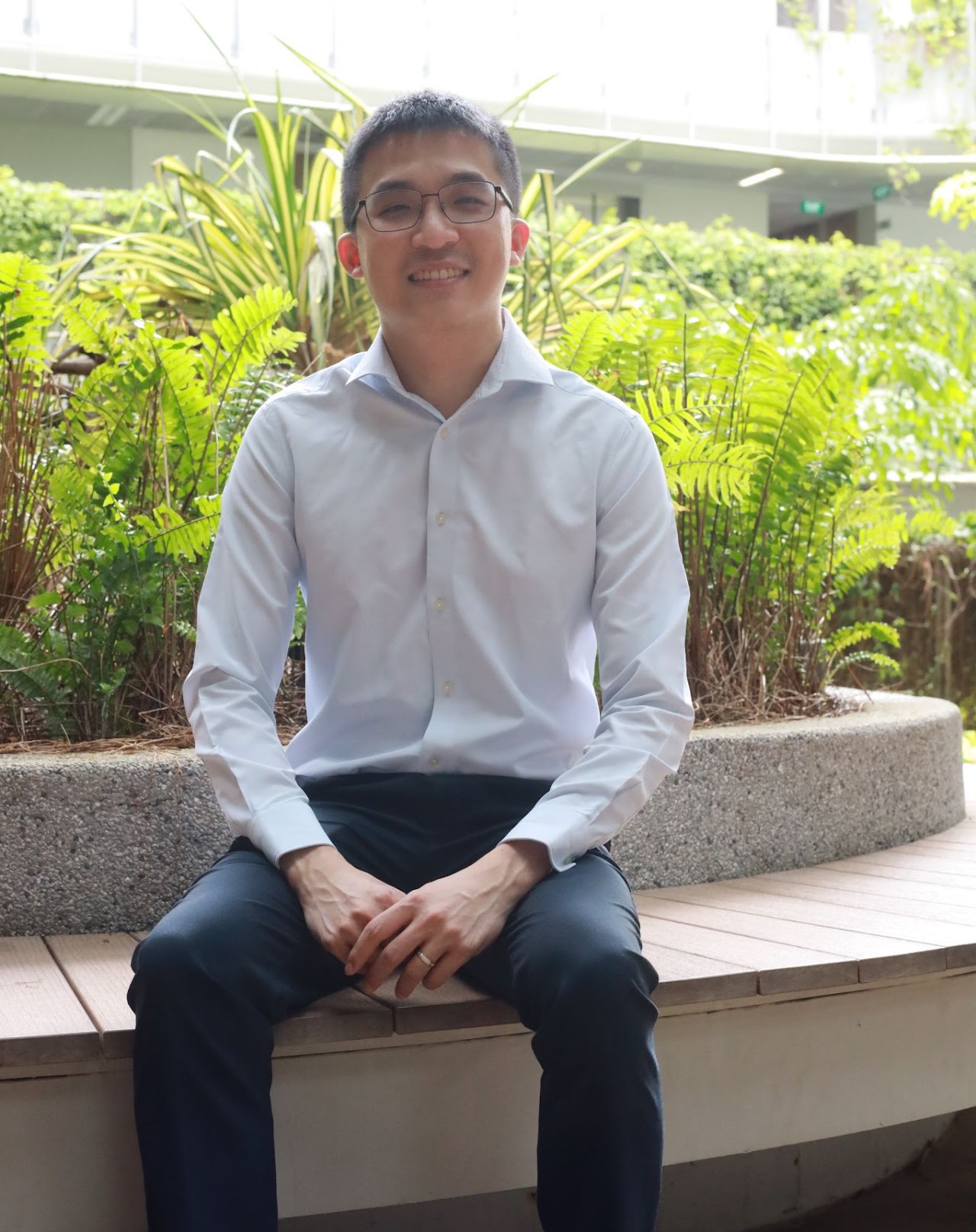}}]{Lim Wei Yang Bryan}
received Ph.D. degree from Nanyang Technological University (NTU) under the Alibaba PhD Talent Programme and was affiliated with the CityBrain team of DAMO academy. He is currently an Assistant Professor at the College of Computing and Data Science (CCDS) in NTU. His doctoral efforts earned him accolades such as the ``Most Promising Industrial Postgraduate Programme Student" award. He also serves on the Technical Programme Committee for FL workshops at flagship conferences (AAAI-FL, IJCAI-FL) and is a review board member for reputable journals like the IEEE TPDS. His research interests include edge intelligence, federated learning, and applied AI.
\end{IEEEbiography}

\end{document}